\documentclass[a4paper,fleqn]{cas-dc}
\usepackage[numbers]{natbib}
\usepackage{framed,multirow}

\usepackage{amssymb}
\usepackage{latexsym}

\usepackage{url}
\usepackage{xcolor}

\usepackage{hyperref}
\usepackage{bm}
\usepackage{url}
\usepackage{multirow}
\usepackage{ulem}
\usepackage{color}

\usepackage{amsmath,amssymb,amsfonts}
\usepackage{algorithmic}
\usepackage{graphicx}
\usepackage{textcomp}
\usepackage{bm}
\usepackage{url}
\usepackage{multirow}
\usepackage{ulem}
\usepackage{color}
\usepackage{pifont}
\usepackage{balance}
\usepackage{colortbl}
\usepackage{xcolor}
\usepackage{makecell}
\usepackage{verbatim}
\newcommand{\cmark}{\ding{51}}

\definecolor{newcolor}{rgb}{.8,.349,.1}
\def\tsc#1{\csdef{#1}{\textsc{\lowercase{#1}}\xspace}}
\tsc{WGM}
\tsc{QE}

\begin{document}
\let\WriteBookmarks\relax
\def\floatpagepagefraction{1}
\def\textpagefraction{.001}
\let\printorcid\relax 

\shorttitle{}    

\shortauthors{Shuwei Shao et al.}

\title[mode = title]{Learnable Patchmatch and Self-Teaching for Multi-Frame Depth Estimation in Monocular Endoscopy }  

\author[1]{Shuwei Shao}
\author[1]{Zhongcai Pei}
\author[1]{Weihai Chen}
\cormark[1]
\author[1]{Xingming Wu}
\author[1]{Zhong Liu}

\address[1]{School of Automation Science and Electrical Engineering, Beihang University, Beijing, China} 
\cortext[1]{Corresponding author}  

\begin{abstract}
	This work delves into unsupervised monocular depth estimation in endoscopy, which leverages adjacent frames to establish a supervisory signal during the training phase. For many clinical applications,~\textit{e.g.}, surgical navigation, temporally correlated frames are also available at test time. Due to the lack of depth clues, making full use of the temporal correlation among multiple video frames at both phases is crucial for accurate depth estimation. However, several challenges in endoscopic scenes, such as low and homogeneous textures and inter-frame brightness fluctuations, limit the performance gain from the temporal correlation. To fully exploit it, we propose a novel unsupervised multi-frame monocular depth estimation model. The proposed model integrates a learnable patchmatch module to adaptively increase the discriminative ability in regions with low and homogeneous textures, and enforces cross-teaching and self-teaching consistencies to provide efficacious regularizations towards brightness fluctuations. Furthermore, as a byproduct of the self-teaching paradigm, the proposed model is able to improve the depth predictions when more frames are input at test time. 
We conduct detailed experiments on multiple datasets, including SCARED, EndoSLAM, Hamlyn and SERV-CT. The experimental results indicate that our model exceeds the state-of-the-art competitors. The source code and trained models will be publicly available upon the acceptance. 
\end{abstract}




\maketitle

	\section{Introduction}
\label{sec:introduction}

Monocular depth estimation has shown to be a practical and versatile technology with a variety of applications, such as surgical navigation~\citep{sinha2018endoscopic} and augmented reality~\citep{chen2018slam}.
While hardware sensors, for instance, structured light, are capable of capturing depth range, the specialist hardware fails to collect dense depth maps and using only a single RGB camera achieves a lower cost. Traditional methods have been explored to estimate depth from monocular videos without using additional hardware sensors such as simultaneously localization and mapping (SLAM)~\citep{grasa2013visual} and structure from motion (SfM)~\citep{leonard2018evaluation}. As a promising alternative, recent learning-based methods leverage adjacent frames to establish a supervisory signal, which eliminates the need of requiring costly hardware sensors to acquire training depth data~\citep{turan2018unsupervised,liu2019dense,li2020unsupervised,shao2021selfsupervised}. In most real-world scenarios, such as surgical navigation, more than one frame is available at both training and test time, and fully exploiting the temporal correlation among multiple frames in these two phases is critical for accurate depth estimation. 

\begin{figure}[!htb]
	\centering
	\includegraphics[width=0.95\linewidth]{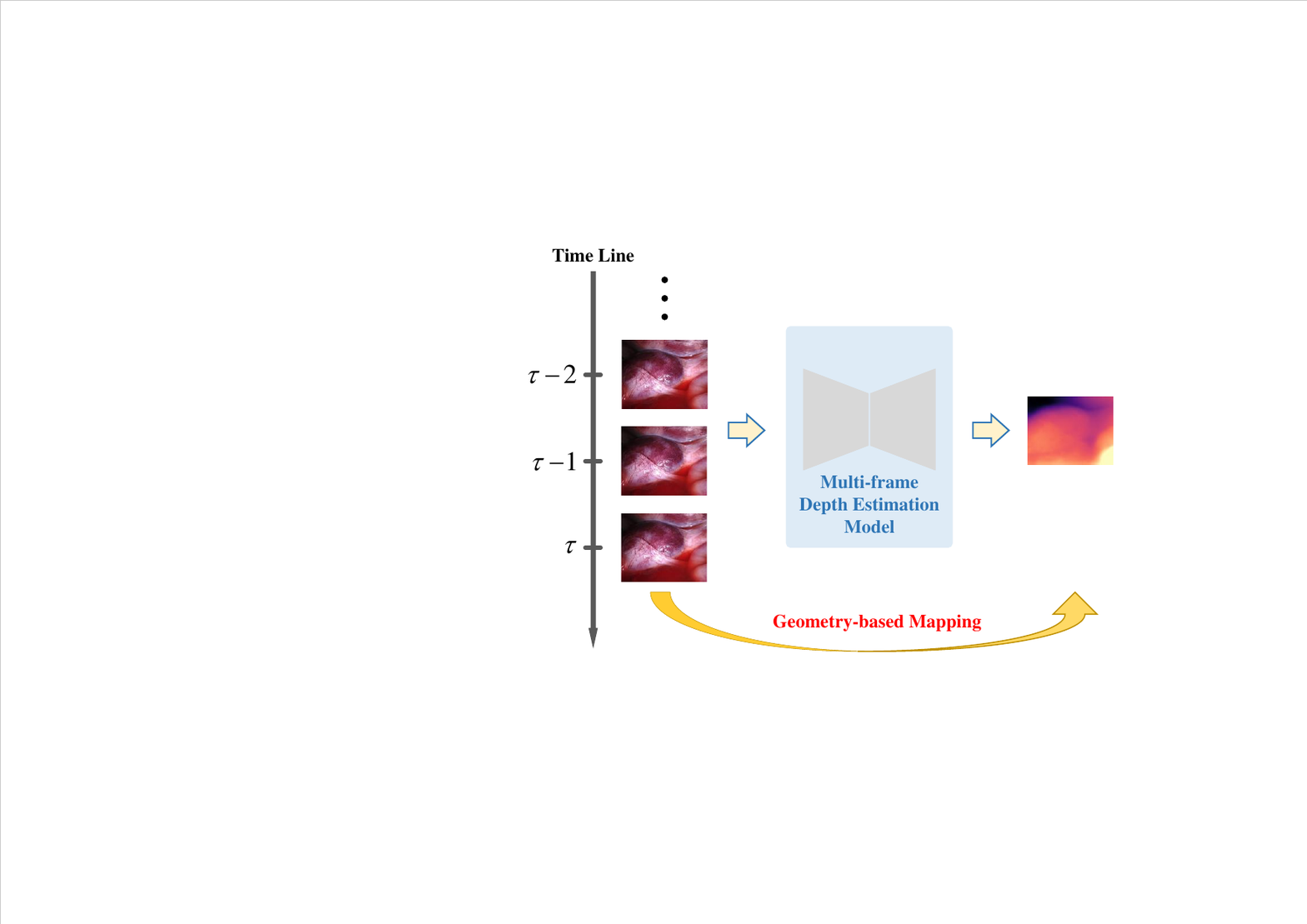}
	\caption{Our method, which trains and tests on endoscopic video streams rather than isolated images, produces accurate depth predictions.}	
	\label{Fig11}
\end{figure}
\begin{figure}[!htb]
	\centering
	\includegraphics[width=0.87\linewidth]{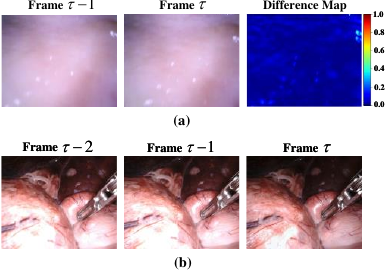}
	\caption{Main challenges encountered in endoscopic scenes. (a) Low texture. The difference map is acquired by taking an absolute difference between frames $\tau$ and $\tau  - 1$, to indicate the low-texture regions. (b) Inter-frame brightness fluctuations. }	
	\label{Fig12}
\end{figure}

Unfortunately, based on our observations and experiments, several challenges in endoscopic scenes limit the performance gain from the temporal correlation, as shown in Fig.~\ref{Fig12}. First, the overall scarce and homogeneous textures of tissues are observed in endoscopy, which makes it hard for the model to obtain reliable information from the photometric loss\footnote{The photometric loss is a main supervisory signal in unsupervised depth estimation.}. For example, the predicted depth error can be very large, while the photometric loss still displays a low value. Second, the brightness constancy assumption is invalid for endoscopy. 
The inter-frame illumination of the same anatomy may vary substantially when the camera and light source move through the environments. In addition, strong non-Lambertian reflections and inter-reflections are caused on the surfaces of smooth tissues and organ fluids. In such regions, the photometric loss and cost volume\footnote{The cost volume is constructed to leverage inputs from multiple views.} are susceptible to be confused by the severe brightness fluctuations. 

To overcome these challenges, we introduce: (i) a \textbf{learnable patchmatch module} to induce a more distinctive characterization in low- and homogeneous-texture regions via adaptive propagation; (ii) \textbf{cross-teaching and self-teaching consistencies} to enhance the robustness towards brightness fluctuations. 

To increase the discriminative ability in regions with low and homogeneous textures, some previous methods leverage the patch-based photometric loss~\citep{furukawa2009accurate,schonberger2016pixelwise,yu2020p}, where a pixel is represented using a local patch, based on the assumption that the pixels within the patch have the same depth. The patch pattern is static,~\textit{e.g.}, a $3\times 3$ grid is used in~\citep{yu2020p}. However, this assumption does not always hold, particularly at boundaries and sharp edge areas due to the abrupt changes in depth. Differently, we propose to aggregate pixels within the local patch in an adaptive manner, taking inspiration from the deformable convolutional networks~\citep{dai2017deformable}. The adaptive propagation is achieved by adding 2D offsets to the regular patch sampling locations. The offsets are learned from a constructed self-correlation volume through additional convolution layers, considering that the pixels with high similarity tend to have the same depth. 
Furthermore, unlike previous methods that only focus on the depth of the patch center pixel,~\textit{e.g.},~\citep{yu2020p}, we use every depth of pixels in the patch to project the center pixel, allowing our learnable patch-based photometric loss to focus on the depth of each pixel in the patch. 

Knowledge distillation (KD) is an efficacious paradigm that transfers learned knowledge from a teacher model to a student one~\citep{hinton2015distilling}. Since then, numerous variants of KD have been proposed using intermediate feature~\citep{heo2019comprehensive}, attention~\citep{komodakis2017paying,hou2019learning}, relation~\citep{yim2017gift,park2019relational},~\textit{etc}. Several studies have also applied KD to monocular depth estimation, using the whole network to distill sub-networks~\citep{pilzer2019refine}, selecting the optimal disparity map from all output scales to distill the whole network~\citep{peng2021excavating}, or using a stereo network to distill a monocular network~\citep{chen2021revealing}. On the other hand, we make the observation that AF-SfMLearner~\citep{shao2021selfsupervised} accounts for variations in brightness patterns by introducing appearance flow, achieving
far less severe errors in brightness fluctuation regions. We leverage the AF-SfMLearner to help teach our model the right answer when brightness fluctuations occur, which we name cross-teaching. 


Data augmentation is a powerful tool for improving training. Recent methods integrate augmentation into learning frameworks via contrastive training, classification~\citep{xie2020unsupervised}, optical flow estimation~\citep{liu2020learning} and hand pose estimation~\citep{yang2021semihand}, to constrain the model predictions to be input noise invariant. 
Intuitively, we design
an appearance simulator to model the edge cases in endoscopic scenes,~\textit{e.g.}, brightness fluctuations and occlusions. Then, we use the frames generated by the appearance simulator and the original frames, respectively, to build cost volumes and enforce the two resulting depth maps to be consistent with each other. The paradigm assists our model in ignoring the detrimental parts in the cost volume and focusing instead on the valuable elements, which we name self-teaching. 


To summarize, the main contributions are listed as follows:

\begin{itemize}
	\item We propose a novel unsupervised multi-frame monocular depth estimation model, where the learnable patchmatch, cross-teaching and self-teaching  provide reliable guidance to make full use of the temporal correlation in endoscopy. Besides, the model can improve the depth predictions when more frames are input at test time. 
	\item We introduce a new learnable patchmatch module to increase the discriminative ability in low- and homogeneous-texture regions via adaptive propagation, which tends to aggregate pixels at the same depth, thus avoiding potential catastrophic errors.
	\item We introduce cross-teaching and self-teaching paradigms to provide efficacious regularizations towards brightness fluctuations. The former is able to strengthen the supervisory signal when brightness changes occur and the latter enables the model to have high immunity against input noise in the cost volume.
	\item Detailed experiments are conducted on SCARED~\citep{allan2021stereo}, EndoSLAM~\citep{bengisu2020quantitative}, Hamlyn\footnote{\url{http://hamlyn.doc.ic.ac.uk/vision/}} and SERV-CT~\citep{edwards2020serv} datasets, which indicate that the proposed model outperforms the state-of-the-art competitors.
\end{itemize}

\begin{figure*}[!htb]
	\centering
	\includegraphics[width=0.99\linewidth]{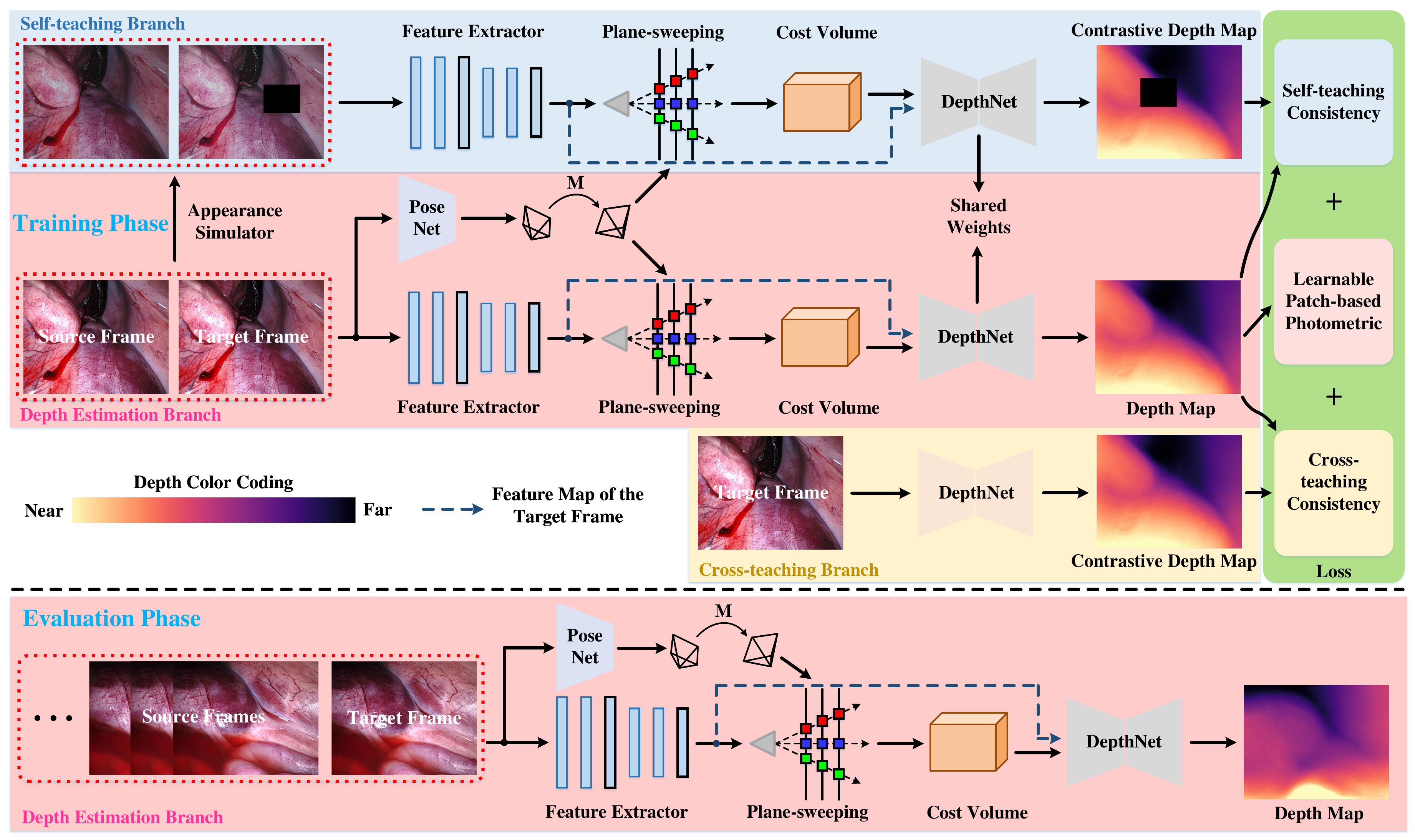}
	\caption{Overview of the whole framework. In the training phase, our framework leverages a target frame and a source frame to build cost volume, and includes three branches, namely self-teaching, depth estimation and cross-teaching. During the evaluation phase, the depth estimation branch is used to produce final results, which allows input of more frames to improve the depth predictions.}
	\label{Fig6}
\end{figure*}

\section{Related work}
\label{sec:related work}
\subsection{Single-frame monocular depth estimation}
Single-frame monocular depth estimation attempts to predict per-pixel depth from an input image. Supervised methods typically use depth supervision from specialist hardwares~\citep{eigen2014depth,cao2017estimating} or synthetic data~\citep{visentini2017deep,mahmood2018deep,mahmood2018unsupervised}. To eliminate the need for expensive ground-truth depth supervision, unsupervised approaches are trained with photometric loss using monocular videos~\citep{zhou2017unsupervised} or stereo pairs~\citep{godard2017unsupervised}.

Recently, significant progress has been made for unsupervised single-frame monocular depth estimation in endoscopy.~\citet{turan2018unsupervised} introduced one of the first studies where they applied SfMLearner~\citep{zhou2017unsupervised} to ex-vivo porcine stomach videos. After that,~\citet{liu2019dense} proposed the use of sparse depths and camera poses from traditional SfM pipelines to establish a supervisory signal.~\citet{bengisu2020quantitative} developed an affine brightness transformer to align the inter-frame brightness condition and a spatial attention module to encourage the pose network to emphasize on highly textured regions.~\citet{shao2021self} introduced appearance flow to take into  variations in brightness patterns. In addition, they adopted a feature scaling module to alleviate the inadequate representation learning issue incurred by low and homogeneous textures. 
Later,~\citet{shao2021selfsupervised} generalized the brightness constancy assumption to a dynamic image constraint, which allows a comprehensive information representation from frame to frame. Yang~\textit{et al.}~\cite{yang2024self1,yang2024self} further developed a lightweight framework to improve efficiency and employed CLIP-guidance segmentation to boost performance, respectively. In contrast to the aforementioned methods, our work focuses on exploiting the temporal correlation during both training and evaluating phases. 
\subsection{Multi-frame monocular depth estimation}
A growing body of studies extend prior single-frame monocular models so that the models can leverage temporal information of video streams to improve the quality of depth predictions at test time. 

~\citet{wang2019recurrent}, ~\citet{zhang2019exploiting} and ~\citet{patil2020don} suggested an idea of combining conventional monocular systems with recurrent networks to process sequences of frames. Similarly, several methods adopted pairs of sequential frames. ~\citet{wang2020self} shared features of depth and pose networks.~\citet{li2020unsupervised} utilized the current frame together with the depth map from the previous frame as the input of depth network. These methods depend heavily on the temporal representations extracted by the network and do not explicitly reason about the inherent geometry. 

~\citet{liu2019neural} and ~\citet{wimbauer2021monorec} integrated a plane-sweep stereo cost volume to predict depth map from multiple frames, which shows significant efficacy in improving results. Unfortunately, they require ground-truth depth supervision to circumvent the scale ambiguity of a monocular system on the construction of cost volume.~\citet{watson2021temporal} introduced an adaptive cost volume that allows its extents to be learned from data, removing the necessity for ground-truth depth. Thereafter, many following-up studies~\cite{wang2023crafting,wang2024multi,wu2023self,xiang2023exploring} have sprung out for urban scenes. In this work, we take one step further and design an efficacious unsupervised multi-frame monocular depth estimation model for endoscopic scenes, which takes into consideration the unique characteristics of minimally invasive surgery environments, such as low and homogeneous textures and inter-frame brightness fluctuations.

\section{Methodology}
\label{method} 
In this section, we first demonstrate preliminary knowledge of unsupervised depth estimation. Next, we construct a cost volume to utilize inputs from multiple views. Then, we introduce learnable patchmatch, cross-teaching and self-teaching to overcome the challenges encountered by an unsupervised multi-frame monocular model in endoscopic scenes. Finally, we present overall architecture and loss.

\subsection{Preliminaries}
Unsupervised depth estimation simulates the learning problem as a task of novel view synthesis, by training a depth network (DepthNet) and a pose network (PoseNet) to synthesize the appearance of a target frame from the view of another frame. The detailed pipeline is described below.

Once the depth of each pixel has been predicted, the pixels are back-projected to a 3D camera space using the depths and the camera intrinsic. Then with the predicted relative pose, the generated point cloud is transformed to another view. Given a target frame ${{{I}}^t}\left( {{\bm{{\rm{p}}}}} \right)$ and a source frame ${{I}^s}\left( {\bm{{\rm{p}}}} \right)$, this process is formulated as
\begin{equation}	
	{{\bm{{\rm{p}}}}^{t \to s}} = {\bm{{\rm{K}}}}{{\bm{{\rm{M}}}}^{t \to s}}{{\bm{{\rm{D}}}}^t}\left( {{\textbf{p}}} \right){{\bm{{\rm{K}}}}^{ - 1}}{{\bm{{\rm{p}}}}^t}, \label{eq1}
\end{equation}	
where $\bm{{\rm{p}}}^{t \to s}$ is the projected pixel coordinates from the target view $t$ to the source view $s$, $\bm{{\rm{p}}}^{t}$ denotes the pixel coordinates in $t$, $\bm{{\rm K}}$ denotes the camera intrinsic, ${\bm{{\rm M}}^{t \to s}}$ denotes the relative pose from $t$ to $s$, and ${\bm{{\rm D}}^t}\left( \bm{{\rm{p}}} \right)$ stands for the depth map of the target frame ${{{I}}^t}\left( {{\bm{{\rm{p}}}}} \right)$. Then we can acquire a synthesized frame
\begin{equation} {I^{s \to t}}\left( {\bm{{\rm{p}}}} \right) = {I^s}\left\langle {{{\bm{{\rm{p}}}}^{t \to s}}} \right\rangle, \label{eq2}
\end{equation}
where $\left\langle  \cdot  \right\rangle$ denotes the warping operation~\citep{jaderberg2015spatial}.
The appearance difference of ${{{I}}^t}\left( {{\bm{{\rm{p}}}}} \right)$ and ${{I}^{s \to t}}\left( {\bm{{\rm{p}}}} \right)$ is delivered as a supervisory signal for the entire training pipeline. Because the supervisory signal generation does not depend on the ground-truth depth, it is called unsupervised depth learning.

The standard practice~\citep{godard2017unsupervised} to quantify appearance difference is to adopt a weighted L1 loss and structural similarity (SSIM) term~\citep{wang2004image}, referred to as photometric loss 
\begin{equation}\begin{split} {{\mathcal{L}}_{ph}}=\sum_{\textbf{p}}\alpha \displaystyle{\frac{1 - {\rm{SSIM}}\left( {{{I^{t}}\left( {\bm{{\rm{p}}}} \right)},{{I^{s \to t}}\left( {\bm{{\rm{p}}}} \right)}} \right)}{2}}\\ +\sum_{\textbf{p}}\left( 1-\alpha  \right){{\left\| {{{I^{t}}\left( {\bm{{\rm{p}}}} \right)} - {{I^{s \to t}}\left( {\bm{{\rm{p}}}} \right)}} \right\|}_{1}}, \end{split} \label{eq3} \end{equation} 
where $\alpha$ is the weight coefficient and is set to 0.85 based on ~\citep{godard2017unsupervised}.
 \begin{figure}[!htb]
	\centering
	\includegraphics[width=0.78\linewidth]{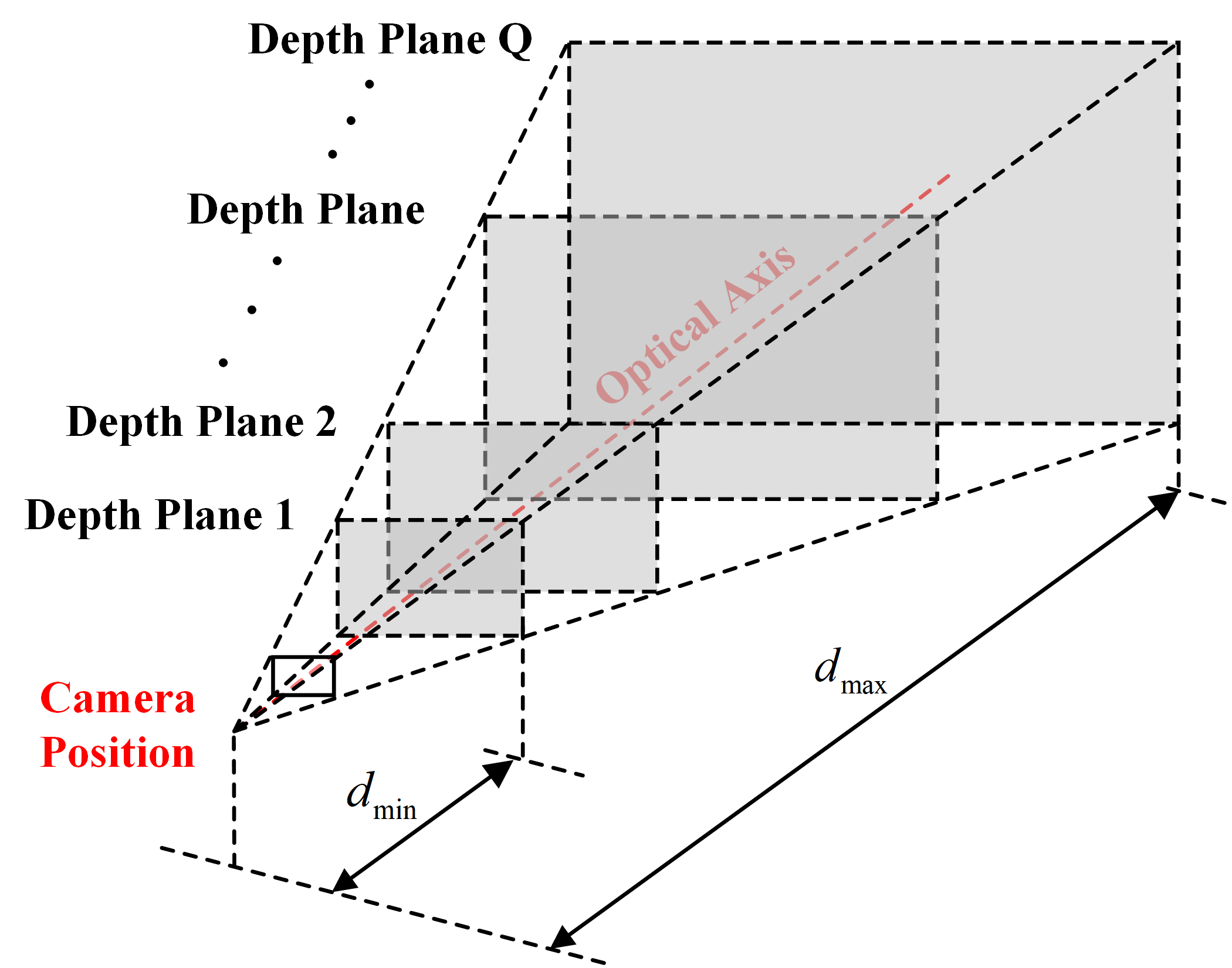}
	\caption{3D Illustration of the depth planes in plane-sweeping, with each plane perpendicular to the optical axis.}	
	\label{Fig10}
\end{figure}


\subsection{Cost volume}
To utilize multiple input frames, we construct a plane-sweep stereo cost volume~\citep{collins1996space} upon the target camera frustum, which measures the geometric coherence of pixels from ${{{I}}^t}\left( {{\bm{{\rm{p}}}}} \right)$ and source frames at different depth values. For the first time we specify a group of fronto-parallel depth planes linearly spaced between the depth values ${d_{\min }}$ and ${d_{\max }}$, as shown in Fig.~\ref{Fig10}. 
Then, for each source frame, we extract a deep feature map and project it to the target camera space using each of the candidate depths, the camera intrinsic and the predicted relative pose by PoseNet. The final cost volume is
constructed as the average of L1 distances between the warped
feature maps and the feature maps from ${{{I}}^t}\left( {{\bm{{\rm{p}}}}} \right)$ over all source frames. When not otherwise specified, we use one source frame in the construction of cost volume during training. The cost volume and feature maps of the target frame are concatenated as the input to DepthNet for depth regression, following~\citep{dosovitskiy2015flownet,watson2021temporal}. 

The cost volume allows a monocular framework to leverage multiple input frames. However, it requires hyperparameters ${d_{\min }}$ and ${d_{\max }}$ to be known. We are unlikely to achieve this because an unsupervised monocular depth estimation model suffers from the scale ambiguity. To relax this constraint, we make ${d_{\min }}$ and ${d_{\max }}$ to be learned from data. In each training iteration, the average min and max of depth predictions over a batch are used to update ${d_{\min }}$ and ${d_{\max }}$ using an exponential moving average with a momentum of 0.99, following~\citep{watson2021temporal}. The updated ${d_{\min }}$ and ${d_{\max }}$ are saved along with the model weights and remain fixed during the evaluation phase.

\begin{figure}[!htb]
	\centering
	\includegraphics[width=0.9\linewidth]{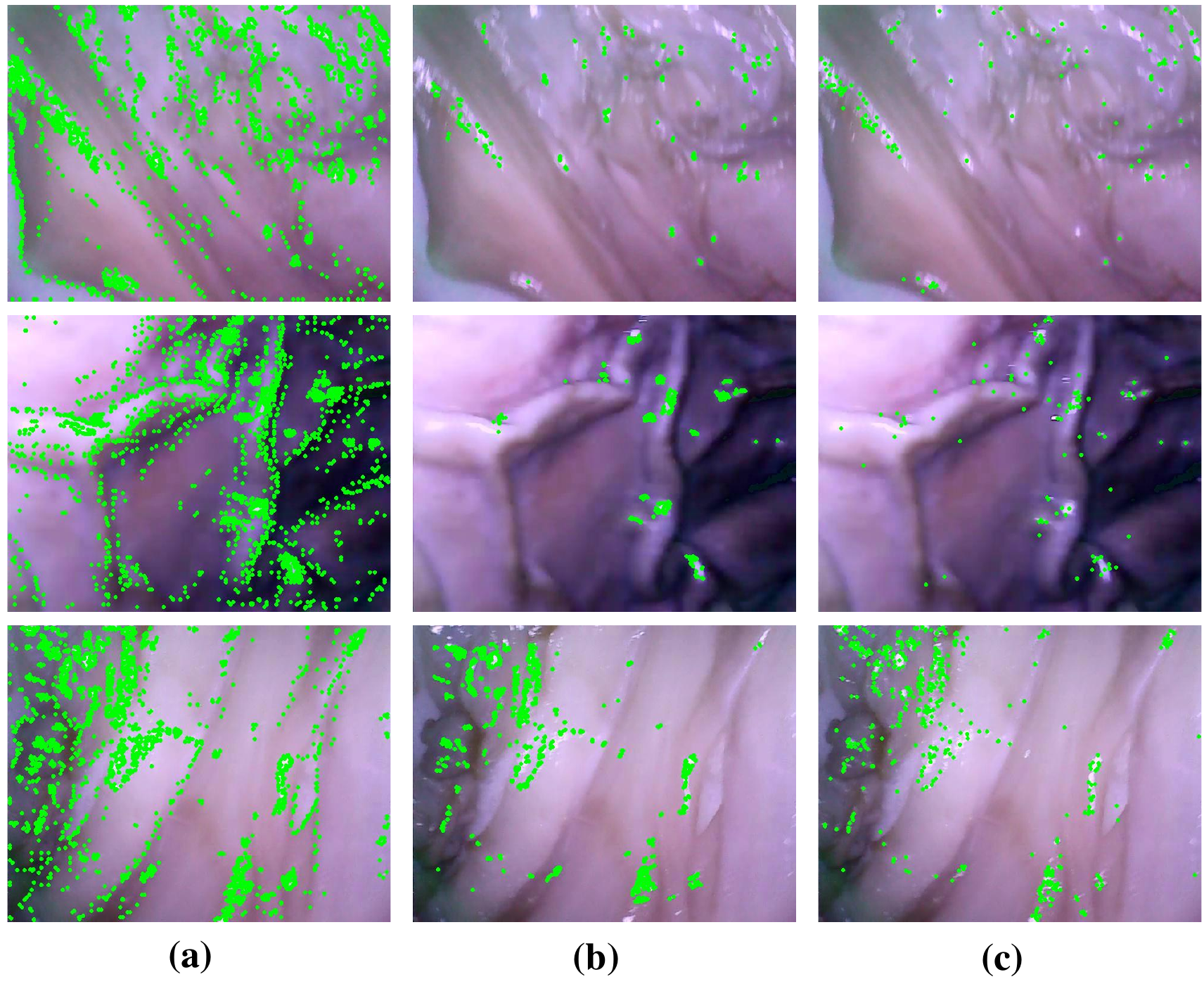}
	\caption{Illustration of the keypoints, which are extracted by algorithms of (a) direct sparse odometry (DSO)~\citep{engel2017direct},  (b) scale-invariant feature transform (SIFT)~\citep{lowe2004distinctive} and (c) oriented fast and rotated brief (ORB)~\citep{rublee2011orb}, and are marked with green circles. The keypoints of DSO are more even and dense.}	
	\label{Fig9}
\end{figure}

\subsection{Learnable patchmatch}
To handle the large low- and homogeneous-texture regions, we introduce a \textbf{learnable patchmatch module}, the details of which are presented below. 

\subsubsection{Keypoints extraction}Following~\citep{yu2020p}, we select representative keypoints as the center pixels in patchmatch. To extract keypoints $\textbf{p}_{k}$, we utilize Direct Sparse Odometry (DSO)~\citep{engel2017direct} for its effectiveness. See Fig.~\ref{Fig9} for an illustration. Note that the extraction of keypoint is performed on-the-fly by DSO, so it can be seamlessly integrated into the training process. We combine each keypoint with a local patch, which is referred to as support domain $\bm{{\rm{\Omega}}}_{\bm{{\rm{p}}}_{k}}$ and relies on the assumption that the pixels within $\bm{{\rm{\Omega}}}_{\bm{{\rm{p}}}_{k}}$ have the same depth. 

\subsubsection{Adaptive propagation}
\begin{figure}[!htb]
	\centering
	\includegraphics[width=1.0\linewidth]{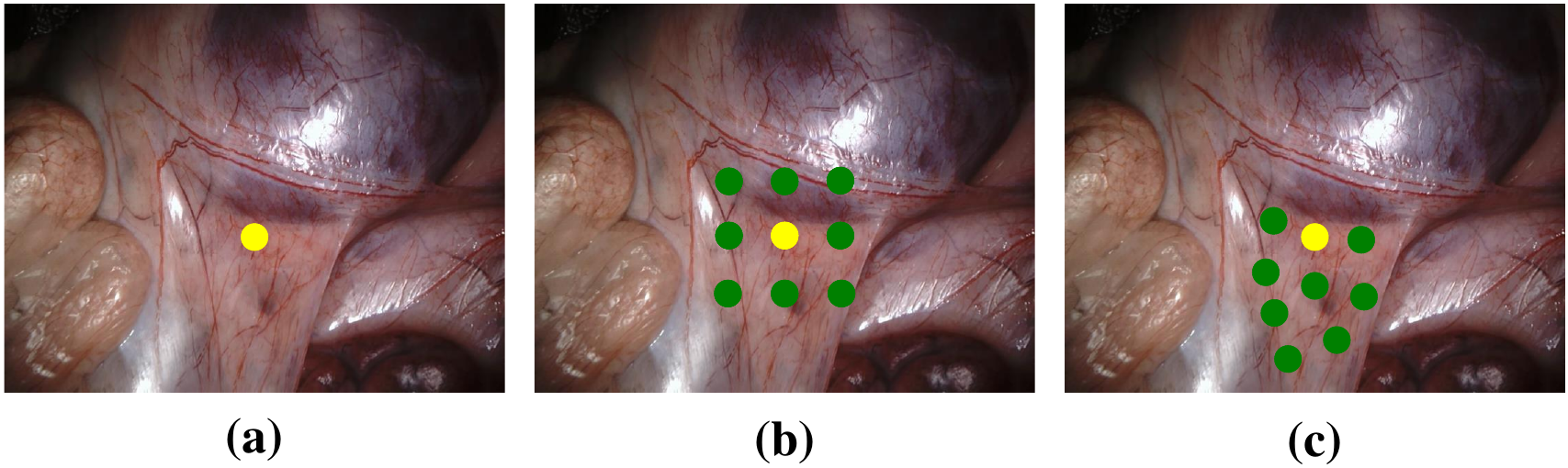}
	\caption{Illustration of the sampled locations in regular patchmatch and our learnable patchmatch.  (a) Target image. (b) Sampled locations of fixed propagation pattern in regular patchmatch. (c) Sampled locations of adaptive propagation pattern in our approach. The yellow marker denotes the center pixel and the green marker denotes the sampled pixel.}	
	\label{Fig8}
\end{figure}

Depth spatial consistency inside a local patch is not always satisfied, such as, at boundaries and sharp edge regions. Hence, instead of adopting a static set of neighbors, such as the $3\times 3$ grid in~\citep{yu2020p}, we propose sampling neighborhood pixels via adaptive propagation. Fig.~\ref{Fig8} depicts the functionality of our adaptive scheme, which tends to aggregate pixels from the same surface, avoiding the catastrophic errors induced by using the fixed pattern.

\begin{figure}[!htb]
	\centering
	\includegraphics[width=1.0\linewidth]{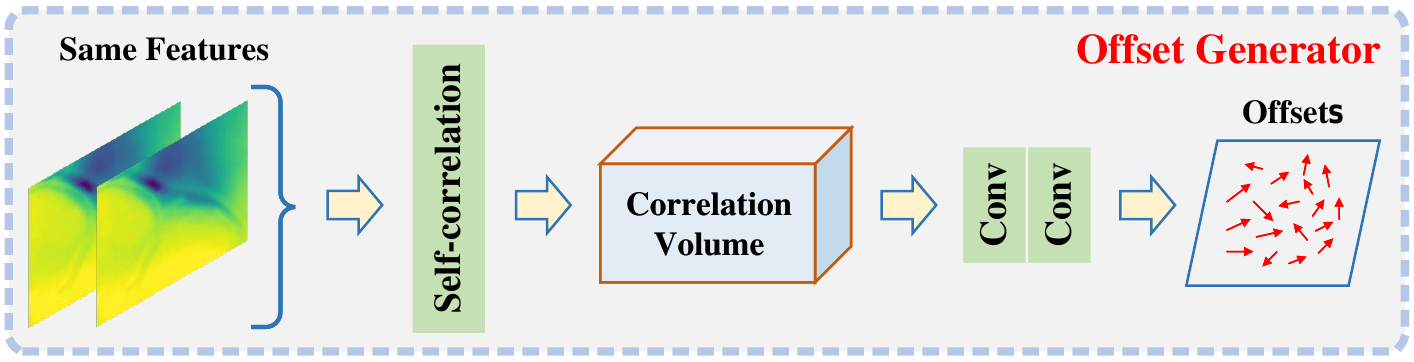}
	\caption{Offset generator. We utilize the penultimate layer feature map of DepthNet to construct the correlation volume for offset decoding. The Conv stands for a $3 \times 3$ convolution layer. The activation function behind the first Conv is ELU~\citep{clevert2015fast} and there is no activation function after the second Conv. }	
	\label{Fig7}
\end{figure}

To adaptively aggregate ${{N}_{{{\textbf{p}}_{k}}}}$ pixels for each keypoints, we learn additional offsets $\left\{ \Delta {{\textbf{o}}_{i}}\left( {{\textbf{p}}_{k}} \right) \right\}_{i=1}^{{{N}_{{{\textbf{p}}_{k}}}}}$ and place the learned offsets on top of a 2D offset grid $\left\{ {\textbf{o}_{i}} \right\}_{i=1}^{{{N}_{{{\textbf{p}}_{k}}}}}$. ${{N}_{{{\textbf{p}}_{k}}}}$ is set to 8 based on~\citep{yu2020p}. As shown in Fig.~\ref{Fig7}, an offset generator is designed considering that the pixels with high similarity are more likely to have the same depth. To be specific, the feature similarity between center keypoint ${{\textbf{p}}_{k}}$ and its surrounding pixels ${{\textbf{p}}_{k}}^{\prime }$ is measured. Mathematically, a correlation index is defined as 
\begin{equation}	
	\textbf{c}\left( {{\textbf{p}}_{k}};\textbf{r} \right)=\mathcal{F}\left( {{\textbf{p}}_{k}} \right)\cdot \mathcal{F}\left( {{\textbf{p}}_{k}}^{\prime } \right)/\ell, \label{eq9}
\end{equation}
where $\textbf{c}$ stands for the correlation vector, $\textbf{r}$ denotes the search range and is set to a $8\times 8$ window (empirical setting), $\mathcal{F}$ denotes the feature map, $\cdot$ denotes the dot product, and $\ell$ denotes the length of feature descriptor. Then, a self-correlation volume is built by integrating all correlation vectors into a 3D grid. Two convolution layers are deployed to decode additional offsets from the correlation volume. With the decoded offsets, we aggregate pixels to constitute the learnable support domain
\begin{equation}
	{\bm{{\rm{\Omega}}}_{{\textbf{p}_{k}}}}=\left\{ {\textbf{p}_{k}}+{\textbf{o}_{i}}+\Delta {\textbf{o}_{i}}\left( {\textbf{p}}_{k} \right) \right\}_{i=1}^{{{N}_{{{\textbf{p}}_{k}}}}}\cup\left\{ {{\textbf{p}_{k}}} \right\}, \label{eq10} 
\end{equation}
where $\cup$ stands for the union of sets. One example of ${\bm{{\rm{\Omega}}}_{{\textbf{p}_{k}}}}$ is presented in Fig.~\ref{Fig8}(c). Subsequently, we acquire the depths of aggregated pixels in ${\bm{{\rm{\Omega}}}_{{\textbf{p}_{k}}}}$ via the warping operation $\left\langle  \cdot  \right\rangle$
\begin{equation}
	{{\textbf{D}}}\left( \bm{{\rm{\Omega}}}_{\bm{{\rm{p}}}_{k}} \right) = \left\{ {{\textbf{D}}}\left\langle {{{\textbf{p}_{k}}+{\textbf{o}_{i}}+\Delta {\textbf{o}_{i}}\left( {\textbf{p}}_{k} \right)}} \right\rangle \right\}_{i=1}^{{{N}_{{{\textbf{p}}_{k}}}}}\cup\left\{ {{\textbf{D}}}\left( {{{\textbf{p}_{k}}}} \right) \right\}.
\end{equation}

\subsubsection{Learnable patch-based photometric loss} Different from traditional view synthesis in unsupervised depth learning, we project the keypoint ${{{\textbf{p}}_{k}}}$ using each depth in the set ${{\textbf{D}}}\left( \bm{{\rm{\Omega}}}_{\bm{{\rm{p}}}_{k}} \right)$. Eq.~\ref{eq1} is thus reformulated as  

\begin{equation}	
	\bm{{\rm{\Omega}}}_{{{\textbf{p}}_{k}}}^{t\to s} = {\bm{{\rm{K}}}}{{\bm{{\rm{M}}}}^{t \to s}}{{\textbf{D}}^{t}}\left( \bm{{\rm{\Omega}}}_{\bm{{\rm{p}}}_{k}} \right){{\bm{{\rm{K}}}}^{ - 1}}{{{\textbf{p}_{k}^{t}}}},  \label{eq6}
\end{equation} 
As such, each $\textbf{p}_{k}^{t}$ is projected into a set $	\bm{{\rm{\Omega}}}_{{{\textbf{p}}_{k}}}^{t\to s}$. Obviously, our patch-based view synthesis differs from~\citep{yu2020p} that utilizes the depth of keypoint ${{{\textbf{p}}_{k}}}$ to project the support domain.
Next, substituting $\bm{{\rm{\Omega}}}_{{{\textbf{p}}_{k}}}$ and $\bm{{\rm{\Omega}}}_{{{\textbf{p}}_{k}}}^{t\to s}$ into Eq.~\ref{eq2}, it can be obtained that  
\begin{equation} {I^{s \to t}}\left( \bm{{\rm{\Omega}}}_{{{\textbf{p}}_{k}}} \right) = {I^s}\left\langle \bm{{\rm{\Omega}}}_{{{\textbf{p}}_{k}}}^{t\to s} \right\rangle. \label{eq7}
\end{equation}
Then, the proposed learnable patch-based photometric loss can be computed as
\begin{equation}\begin{split} {{\mathcal{L}}_{lpph}}=\sum_{\bm{{\rm{\Omega}}}_{{{\textbf{p}}_{k}}}}\alpha \displaystyle{\frac{1 - {\rm{SSIM}}\left( {{{I^{t}}\left( \textbf{p}_k \right)},{{I^{s \to t}}\left( \bm{{\rm{\Omega}}}_{{{\textbf{p}}_{k}}} \right)}} \right)}{2}}\\ +\sum_{\bm{{\rm{\Omega}}}_{{{\textbf{p}}_{k}}}}\left( 1-\alpha  \right){{\left\| {{{I^{t}}\left( \textbf{p}_k \right)} - {{I^{s \to t}}\left( \bm{{\rm{\Omega}}}_{{{\textbf{p}}_{k}}} \right)}} \right\|}_{1}}, \end{split} \label{eq8} \end{equation}  
where each element in ${{I^{s \to t}}\left( \bm{{\rm{\Omega}}}_{{{\textbf{p}}_{k}}} \right)}$ calculates the photometric error with  ${{I^{t}}\left( \textbf{p}_k \right)}$. It is worth noting that the loss ${{\mathcal{L}}_{lpph}}$ not only supervises the depths of keypoints, but also supervises the depths of non-keypoint pixels using a more discriminative supervisory signal at each keypoint. In order to alleviate the negative impacts of occluded pixels on ${{\mathcal{L}}_{lpph}}$, two source frames ${{I}^{s}}\left( \textbf{p} \right)\in \left\{ {{I}^{t-1}}\left( \textbf{p} \right),{{I}^{t+1}}\left( \textbf{p} \right) \right\}$ are used to define the photometric loss, and the one with the minimum ${{\mathcal{L}}_{lpph}}$ is chosen~\citep{godard2019digging}. Note that the source frames used in ${{\mathcal{L}}_{lpph}}$ are not the same as the source frames to build cost volume.

\subsection{Cross-teaching}
To reinforce the supervisory signal when brightness fluctuations occur, we propose a \textbf{cross-teaching paradigm}, based on the observation that  AF-SfMLearner~\citep{shao2021selfsupervised} achieves far less severe errors in the depth of brightness change regions benefiting from the introduced appearance flow. However, adding the appearance flow term into the photometric loss, as in~\citep{shao2021selfsupervised}, leads to expensive memory consumption. We therefore leverage the
trained depth network by the AF-SfMLearner to teach our model towards the correct depth even in areas with large brightness variations, which is more memory efficient. For each training sample, this separate network generates a depth map ${{\widehat{\textbf{D}}}^{t}}\left( \textbf{p} \right)$ and is discarded when the training is completed. Mathematically, our loss function is defined as 
\begin{equation} 
	{{\cal L}_{ct}} =\sum_{\textbf{p}}\frac{\left| {{\textbf{D}}^{t}}\left( \textbf{p} \right)-{{\widehat{\textbf{D}}}^{t}}\left( \textbf{p} \right) \right|}{{{\textbf{D}}^{t}}\left( \textbf{p} \right)+{{\widehat{\textbf{D}}}^{t}}\left( \textbf{p} \right)},
\end{equation} 
where ${{\cal L}_{ct}}$ denotes the cross-teaching consistency loss. Gradients to ${{\widehat{\textbf{D}}}^{t}}\left( \textbf{p} \right)$ are stopped to guarantee that teacher delivers knowledge to student and not vice versa. Here, instead of employing their absolute difference directly, we normalize the absolute difference by their sum. 	
This is more intuitive since pixels with different absolute depths are treated equally during optimization. In addition, the outputs are scale-invariant with a natural range of 0 to 1, which is beneficial to the numerical stability during training.

\subsection{Self-teaching}
A \textbf{self-teaching paradigm} is further developed to help our model to have high immunity against input noise in the cost volume, such as that induced by brightness fluctuations and occlusions, and focusing on the valuable elements. Towards this end, we design an appearance simulator consisting of random gamma correction, color jitter and masking to simulate the edge cases. We use the frames derived by the appearance simulator and the original frames to construct cost volumes, respectively, and enforce the two resulting depth maps to be consistent with each other, which provides robustness against brightness fluctuations and occlusions. The self-teaching consistency loss is defined as 
\begin{equation} 
	{{\cal L}_{st}} =\sum_{\textbf{p}}\frac{\left| {{\textbf{D}}^{t}}\left( \textbf{p} \right)-{{\bar{\textbf{D}}}^{t}}\left( \textbf{p} \right) \right|}{{{\textbf{D}}^{t}}\left( \textbf{p} \right)+{{\bar{\textbf{D}}}^{t}}\left( \textbf{p} \right)}\odot \textbf{R}\left( \textbf{p} \right),
\end{equation}
where ${{\bar{\textbf{D}}}^{t}}\left( \textbf{p} \right)$ represents the depth map of transformed frames, $\odot $ denotes the element-wise multiplication, and $\textbf{R}\left( \textbf{p} \right)$ denotes the unoccluded mask, determined by random masking. Gradients to ${{\textbf{D}}^{t}}\left( \textbf{p} \right)$ are stopped. We describe the components of appearance simulator in detail below including random gamma correction, color jitter and masking.

\textbf{Gamma correction} is a non-linear mapping used to adjust the illuminance of images. During the data acquisition process of the endoscopic scenes, there are severe brightness fluctuations caused by illumination variations. To simulate the varied illuminations, we incorporate random gamma correction.

\textbf{Color jitter} mainly involves jitters of lightness, saturation, hue and contrast. We integrate random color jitter to mimic the complicated non-Lambertian reflections and inter-reflections in minimally invasive surgery environments.

\textbf{Masking} is intended to mimic the occlusions among multiple input frames. To be specific, we generate a binary mask to randomly crop some portions on target frame, and then adopt the binary mask as a ground-truth of the mimicked occlusion in self-teaching consistency loss, which contributes to make the model insensitive to occlusion, that is, even when there is occlusion, the model can still correctly predict the depth of the unoccluded areas.

\subsection{Overall architecture and loss}
\subsubsection{Overall architecture}Fig.~\ref{Fig6} presents an overview of the proposed framework, which involves three parts: depth estimation branch, cross-teaching branch and self-teaching branch. The feature extractor contains the first five ResNet18 layers~\citep{he2016deep}. The DepthNet adopts an encoder-decoder structure with skip connections, where the remaining ResNet18 convolution layers are used as the encoder and the decoder shares similar structure to~\citep{watson2021temporal}. The PoseNet is the same as~\citep{godard2019digging}. 

\begin{table*}[htb!]
	\caption{Quantitative depth comparison on the SCARED dataset. V denotes the number of input views. CIs denotes the confidence intervals. The top two results are highlighted in \textcolor{red}{red} and \textcolor{blue}{blue}. ``-'' means not applicable.}
	\begin{center}
		\renewcommand{\arraystretch}{1.3}
		\resizebox{2.0\columnwidth}{!}{\begin{tabular}{c c c c c c c c c c c c}	
				\Xhline{1.2pt}
				Method & V & \cellcolor {orange!20} Abs Rel $\downarrow$ & 95\% CIs & \cellcolor {orange!20} Sq Rel $\downarrow$ & 95\% CIs & \cellcolor {orange!20} RMSE $\downarrow$& 95\% CIs& \cellcolor {orange!20} RMSE log $\downarrow$ &95\% CIs & \cellcolor {blue!20} $\delta$ $\uparrow$ & 95\% CIs\\ 
				\hline						
				\hline
				SfMLearner~\citep{zhou2017unsupervised} &1& 0.079 & [0.076, 0.081] & 0.879  & [0.794, 0.964] & 6.896 & [6.513, 7.279] & 0.110 & [0.106, 0.115]& 0.947 & [0.942, 0.952]\\
				DeFeat-Net~\citep{spencer2020defeat} &1& 0.077 & [0.074, 0.079] & 0.792 & [0.731, 0.853] & 6.688 & [6.355, 7.021]& 0.108 & [0.104, 0.112] & 0.941 & [0.936, 0.946] \\
				Monodepth2~\citep{godard2019digging} &1& 0.071 & [0.068, 0.073] & 0.590 & [0.554, 0.627]& 5.606& [5.404, 5.809] & 0.094 & [0.091, 0.097] & 0.953 & [0.948, 0.957]\\
				Endo-SfM~\citep{bengisu2020quantitative}&1&  0.062 &[0.060, 0.065] & 0.606 & [0.551, 0.661] & 5.726 & [5.396, 6.056]& 0.093 & [0.089, 0.097]& 0.957 & [0.952, 0.961]\\
				Yang~\textit{et al.}~\cite{yang2024self1}&1& 0.062& -& 0.558 & -& 5.585& -& 0.090 & -& 0.962& -\\
				AF-SfMLearner~\citep{shao2021selfsupervised} &1& \textcolor{blue}{0.059}& [0.057, 0.061]& 0.435 & [0.406, 0.464]& 4.925& [4.729, 5.122]& 0.082 & [0.079, 0.084]& 0.974& [0.971, 0.977]\\
				\hline
				\hline
				ManyDepth~\cite{watson2021temporal} & 2 & 0.061&[0.059, 0.063] &0.505 &[0.466, 0.543] & 5.371&[5.112, 5.631]&0.087&[0.084, 0.090]&0.967&[0.964, 0.971]
				\\
				\textbf{Ours} &2 & \textcolor{red}{0.051}& [0.050, 0.053]&0.386 &[0.354, 0.417] & 4.675&[4.451, 4.899]&\textcolor{blue}{0.074}&[0.072, 0.077]&\textcolor{blue}{0.979}&[0.977, 0.982]
				\\
				\hline
				ManyDepth~\cite{watson2021temporal} &3  &0.062&[0.060, 0.064] &0.511 &[0.473, 0.549] &5.415 &[5.160, 5.669]&0.088&[0.085, 0.091]&0.966&[0.963, 0.970]
				\\
				\textbf{Ours} & 3 & \textcolor{red}{0.051}&[0.049, 0.052] &0.374&[0.344, 0.405]&4.615 &[4.393, 4.838]&\textcolor{red}{0.073}&[0.071, 0.076]&\textcolor{red}{0.980}&[0.977, 0.982]
				\\
				\hline
				ManyDepth~\cite{watson2021temporal} & 4  &0.064 &[0.062, 0.066] &0.527& [0.489, 0.565]&5.494 &[5.243, 5.745]&0.090&[0.087, 0.094]&0.964&[0.960, 0.968]
				\\
				\textbf{Ours} &4 &\textcolor{red}{0.051} &[0.049, 0.052]&\textcolor{blue}{0.370}&[0.340, 0.400]&\textcolor{blue}{4.600} &[4.380, 4.820]&\textcolor{red}{0.073}&[0.071, 0.076]&\textcolor{red}{0.980}&[0.977, 0.983]
				\\
				\hline
				ManyDepth~\cite{watson2021temporal} & 5  &0.066&[0.064, 0.068] & 0.544&[0.506, 0.583] &5.572 &[5.323, 5.821]&0.092&[0.089, 0.096]&0.961&[0.957, 0.965]
				\\
				\textbf{Ours} & 5 & \textcolor{red}{0.051}&[0.049, 0.053] & \textcolor{red}{0.368}&[0.338, 0.397] &\textcolor{red}{4.597}&[4.380, 4.814]&\textcolor{red}{0.073}&[0.071, 0.076]&\textcolor{red}{0.980} &[0.977, 0.983]
				\\
				\Xhline{1.2pt}
		\end{tabular}}
	\end{center}
	\label{table1}
\end{table*}

\begin{table*}[htb!]
	\caption{Point cloud comparison on the SCARED dataset. }
	\begin{center}
		\renewcommand{\arraystretch}{1.3}
		\resizebox{2.0\columnwidth}{!}{\begin{tabular}{c c c c c c c c c c c c}	
				\Xhline{1.2pt}
				Method & V& \cellcolor {orange!20} Abs Rel $\downarrow$ & 95\% CIs & \cellcolor {orange!20} Sq Rel $\downarrow$ & 95\% CIs & \cellcolor {orange!20} RMSE $\downarrow$& 95\% CIs& \cellcolor {orange!20} RMSE log $\downarrow$ &95\% CIs & \cellcolor {orange!20} MSE $\downarrow$ & 95\% CIs\\ 
				\hline						
				\hline
				SfMLearner~\citep{zhou2017unsupervised} &1& 0.087 & [0.084, 0.091] & 0.480  & [0.431, 0.528] & 4.446 & [4.192, 4.701] & 0.120 & [0.116, 0.124]& 29.021 & [25.424, 32.617]\\
				DeFeat-Net~\citep{spencer2020defeat} &1& 0.085 & [0.081, 0.089] & 0.424 & [0.390, 0.457] & 4.302 & [4.083, 4.520]& 0.118 & [0.114, 0.122] & 25.345 & [22.597, 28.092] \\
				Monodepth2~\citep{godard2019digging} &1& 0.079 & [0.076, 0.083] & 0.319 & [0.299, 0.338]& 3.614& [3.482, 3.747] &0.105 & [0.102, 0.108] & 15.567 & [14.459, 16.675]\\
				Endo-SfM~\citep{bengisu2020quantitative}&1&  0.071 &[0.067, 0.074] & 0.331 & [0.299, 0.362] & 3.702 & [3.483, 3.920]& 0.105 & [0.101, 0.108]& 20.521 & [17.846, 23.195]\\
				AF-SfMLearner~\citep{shao2021selfsupervised} &1& 0.068& [0.065, 0.071]& 0.234 & [0.218, 0.249]& 3.174& [3.046, 3.302]& 0.094 & [0.091, 0.096]& 12.429& [11.364, 13.495]\\
				\hline
				\hline
				ManyDepth~\cite{watson2021temporal} & 2 & 0.070&[0.067, 0.073] &0.276 &[0.254, 0.297] & 3.476&[3.303, 3.648]&0.099&[0.096, 0.102]&16.341&[14.512, 18.171]
				\\
				\textbf{Ours} &2 & \textcolor{blue}{0.060}& [0.057, 0.063]&0.209 &[0.192, 0.226] & 3.020&[2.873, 3.167]&\textcolor{blue}{0.087}&[0.085, 0.090]&12.216&[10.873, 13.560]
				\\
				\hline
				ManyDepth~\cite{watson2021temporal} &3  &0.071&[0.068, 0.074] &0.279 &[0.258, 0.300] &3.503 &[3.334, 3.672]&0.100&[0.097, 0.103]&16.363&[14.571, 18.154]
				\\
				\textbf{Ours} & 3 & \textcolor{red}{0.059}&[0.056, 0.062] &0.203&[0.186, 0.220]&2.982 &[2.835, 3.128]&\textcolor{blue}{0.087} &[0.084, 0.089]&11.959&[10.629, 13.288]
				\\
				\hline
				ManyDepth~\cite{watson2021temporal} & 4  &0.073 &[0.069, 0.076] &0.287& [0.266, 0.308]&3.552 &[3.385, 3.719]&0.102&[0.099, 0.104]&16.594&[14.832, 18.357]
				\\
				\textbf{Ours} &4 &\textcolor{red}{0.059} &[0.056, 0.062]&\textcolor{blue}{0.200}&[0.184, 0.217]&\textcolor{blue}{2.971} &[2.827, 3.116]&\textcolor{red}{0.086}&[0.084, 0.089]&\textcolor{blue}{11.821}&[10.519, 13.124]
				\\
				\hline
				ManyDepth~\cite{watson2021temporal} & 5  &0.075&[0.071, 0.078] & 0.296&[0.275, 0.317] &3.600 &[3.435, 3.766]&0.103&[0.100, 0.106]&16.881&[15.131, 18.631]
				\\
				\textbf{Ours} & 5 & \textcolor{blue}{0.060}&[0.057, 0.063] & \textcolor{red}{0.199}&[0.183, 0.215] &\textcolor{red}{2.969}&[2.826, 3.112]&\textcolor{red}{0.086}&[0.084, 0.089]&\textcolor{red}{11.728} &[10.457, 13.000]
				\\
				\Xhline{1.2pt}
		\end{tabular}}
	\end{center}
	\label{table6}
\end{table*}

\begin{table}[htb!]
	\caption{Quantitative pose comparison on the SCARED dataset. Based on~\citep{zhou2017unsupervised}, the ATE is calculated on 5-frame snippets and averaged over the whole sequence. Res-18: ResNet-18.}
	\begin{center}
		\renewcommand{\arraystretch}{1.3}
		\resizebox{1.0\columnwidth}{!}{\begin{tabular}{c c c c c c }
				\hline
				\multirow{2}{*}{Method} & \multirow{2}{*}{Backbone} & \multicolumn{2}{c}{Trajectory-1 }&  \multicolumn{2}{c}{Trajectory-2 } \\ 
				\cline{3-6} 
				&& ATE& 95\% CIs & ATE & 95\% CIs    \\ 
				\hline
				\hline
				DeFeat-Net~\citep{spencer2020defeat}& Res-18 & 0.1765 &[0.1658, 0.1872]& 0.0995&[0.0953, 0.1037]\\
				Monodepth2~\citep{godard2019digging}& Res-18 &
				0.0769&[0.0718, 0.0820]& 0.0554&[0.0532, 0.0576]\\
				Endo-SfM~\citep{bengisu2020quantitative}& Res-18 & 0.0759 & [0.0709, 0.0809]& 0.0500 &[0.0480, 0.0519]\\			
				ManyDepth~\cite{watson2021temporal}& Res-18 & 0.0762 & [0.0712, 0.0812]& 0.0576 &[0.0555, 0.0596]\\		
				AF-SfMLearner~\citep{shao2021selfsupervised}   & Res-18 & 0.0742 &[0.0692, 0.0792]& 0.0478& [0.0459, 0.0497] \\
				Yang~\textit{et al.}~\cite{yang2024self1}& Res-18+MHA2& \textcolor{blue}{0.0723} &-& \textcolor{blue}{0.0474}& - \\
				\textbf{Ours} & Res-18 & \textcolor{red}{0.0719} &[0.0669, 0.0768]& \textcolor{red}{0.0426}& [0.0408, 0.0444] \\	
				\hline
		\end{tabular}}
	\end{center}
	\label{table7}
\end{table}
\subsubsection{Overall loss}Apart from the learnable patch-based photometric loss ${{\mathcal{L}}_{lpph}}$, cross-teaching consistency loss ${{\cal L}_{ct}}$ and self-teaching consistency loss ${{\cal L}_{st}}$,  an edge-aware smoothness loss is utilized to encourage the smoothness property of depth map, which is defined as~\citep{godard2017unsupervised}
\begin{equation}
	{{\cal L}_{es}} =\sum_{\textbf{p}}{\left| { {\nabla {\bf{D} }\left( \bf{p} \right)}} \right|} * {e^{ - \left| \nabla {{I^t}\left( \bf{p} \right)} \right|}}.
\end{equation}
The total optimization objective is defined as follows
\begin{equation}
	{{\cal L}_{total}} = {{\lambda }_{1}}{{\mathcal{L}}_{lpph}} + {{\lambda }_{2}}{{\cal L}_{ct}} + {{\lambda }_{3}}{{\cal L}_{st}} + {{\lambda }_{4}}{{\cal L}_{es}},
\end{equation}
where ${{\lambda }_{1}} = 1$, ${{\lambda }_{2}} = 0.02$, ${{\lambda }_{3}} = 0.002$, ${{\lambda }_{4}} = 0.0001$, and the weights of loss are empirically set to an initial value and then fine-tuned on the SCARED dataset. Since ${{\cal L}_{total}}$ does not rely on the ground-truth depth, we refer to our method an unsupervised depth learning.

		
		

\section{Experiment}

\subsection{Datasets}
\begin{itemize}
	\item \textbf{SCARED}~\citep{allan2021stereo} consists of 35 endoscopic videos collected from porcine cadaver abdominal anatomy, along with the point cloud and pose ground-truths.
	\item \textbf{EndoSLAM}~\citep{allan2021stereo} is collected from  porcine gastrointestinal tract organs and a silicone colon phantom model. It involves challenging real part with the pose ground-truth and synthetic part with the depth and pose ground-truths.
	\item \textbf{Hamlyn}\footnote{\url{http://hamlyn.doc.ic.ac.uk/vision/}} contains phantom heart model videos with the point cloud ground-truth and in vivo endoscopic videos. We utilize the dataset processed by~\citep{recasens2021endo}, where they annotate the in vivo videos using the Efficient Large-Scale Stereo Matching algorithm~\citep{geiger2010efficient}. 
	\item \textbf{SERV-CT}~\citep{edwards2020serv} is composed of 16 stereo pairs acquired from porcine torso cadavers with the depth and disparity ground-truths.
\end{itemize}

\begin{table*}[htb!]
	\caption{Quantitative pose comparison on the EndoSLAM dataset.  }
	\begin{center}
		\smallskip
		\renewcommand{\arraystretch}{1.3}
		\resizebox{2.0\columnwidth}{!}{\begin{tabular}{c c c c c c c c c c c c}
				\hline
				&\multirow{2}{*}{Method} & \multicolumn{2}{c}{Trajectory-1 }&  \multicolumn{2}{c}{Trajectory-2 } & \multicolumn{2}{c}{Trajectory-3 }& \multicolumn{2}{c}{Trajectory-4 }& \multicolumn{2}{c}{Trajectory-5 } \\ 
				\cline{3-12} 
				&& ATE ($\times$ 1e-2)& 95\% CIs & ATE ($\times$ 1e-2) &  95\% CIs& ATE ($\times$ 1e-2) &  95\% CIs & ATE ($\times$ 1e-2) &  95\% CIs & ATE ($\times$ 1e-2) &  95\% CIs    \\ 
				\hline
				\hline
				\multirow{4}{*}{Colon-IV} 
				& Endo-SfM & 0.1996 &[0.0634, 0.3358]& 0.1090&[0.0915, 0.1265] & 0.0828&[0.0727, 0.0928]&  0.0981&[0.0480, 0.1482]& \textcolor{blue}{0.0948}&[0.0838, 0.1059] \\
				& ManyDepth & 0.1549 &[0.0691, 0.2407]& 0.1145&[0.0968, 0.1322] & 0.0961&[0.0857, 0.1064]& 0.1061&[0.0671, 0.1450]& 0.1094&[0.0991, 0.1196] \\
				& AF-SfMLearner &  \textcolor{blue}{0.1297} &[0.0644, 0.1951]&  \textcolor{blue}{0.0946}&[0.0813, 0.1080] &  \textcolor{red}{0.0674}&[0.0612, 0.0737]& \textcolor{blue}{0.0802}&[0.0554, 0.1051]&  0.1002&[0.0908, 0.1095]\\
				& \textbf{Ours} & \textcolor{red}{0.1231} &[0.0575, 0.1888]&  \textcolor{red}{0.0772}&[0.0670, 0.0873] &  \textcolor{blue}{0.0686}&[0.0623, 0.0749]& \textcolor{red}{0.0780}&[0.0524, 0.1036]&  \textcolor{red}{0.0886}&[0.0808, 0.0964]\\
				\hline
				\multirow{4}{*}{Intestine} 
				& Endo-SfM & 0.1212 &[0.1123, 0.1301]& 0.1205 &[0.1081, 0.1330] &  \textcolor{red}{0.0795}&[0.0719, 0.0871]&  \textcolor{red}{0.0663}&[0.0448, 0.0877]& 0.5255 &[0.4876, 0.5633] \\
				& ManyDepth & 0.1443 &[0.1340, 0.1545]& 0.1292&[0.1166, 0.1419] & 0.0953&[0.0884, 0.1023]& 0.0891&[0.0628, 0.1154]& 0.5252&[0.4863, 0.5641] \\
				& AF-SfMLearner &  \textcolor{blue}{0.0894} &[0.0842, 0.0946]&  \textcolor{blue}{0.0826}&[0.0753, 0.0898] & 0.0930 &[0.0873, 0.0987]& 0.0808 &[0.0643, 0.0974]&  \textcolor{blue}{0.2502}&[0.2309, 0.2694] \\
				& \textbf{Ours} &  \textcolor{red}{0.0828} &[0.0781, 0.0874]&  \textcolor{red}{0.0792}&[0.0722, 0.0862] & \textcolor{blue}{0.0925} &[0.0870, 0.0981]& \textcolor{blue}{0.0781} &[0.0661, 0.0900]&  \textcolor{red}{0.2219}&[0.2036, 0.2402] \\
				\hline
				\multirow{4}{*}{Stomach-I} 
				& Endo-SfM & 0.1335 &[0.1137, 0.1533]& 0.1230&[0.1032, 0.1429] & 0.4040&[0.3211, 0.4870]&  \textcolor{red}{0.1893}&[0.0884, 0.2901]& -&- \\
				& ManyDepth & 0.1607 &[0.1372, 0.1841]& 0.1389&[0.1175, 0.1604] & 0.4306&[0.3400, 0.5212]& 0.2331&[0.1138, 0.3524]& -&- \\
				& AF-SfMLearner &  \textcolor{blue}{0.1129} &[0.1002, 0.1256]&  \textcolor{blue}{0.1098}&[0.0929, 0.1267] &  \textcolor{blue}{0.3834}&[0.2885, 0.4782]& 0.2276&[0.1110, 0.3442]& -&- \\
				& \textbf{Ours} &  \textcolor{red}{0.1010} &[0.0930, 0.1090]&  \textcolor{red}{0.0985}&[0.0814, 0.1157] &  \textcolor{red}{0.3297}&[0.2581, 0.4014]& \textcolor{blue}{0.2270}&[0.1057, 0.3483]& -&- \\
				\hline
				\multirow{4}{*}{Stomach-II} 
				& Endo-SfM & 0.1047 &[0.0949, 0.1145]& 0.0514&[0.0225, 0.0803] & 0.1153&[0.1030, 0.1276]& 0.1279 &[0.1156, 0.1402]& -&- \\
				& ManyDepth & 0.1103 &[0.1006, 0.1200]&  0.0578&[0.0229, 0.0927] & 0.1177 &[0.1051, 0.1302]& 0.1206&[0.1094, 0.1318]& -&- \\
				& AF-SfMLearner &  \textcolor{blue}{0.0917} &[0.0836, 0.0999]& \textcolor{blue}{0.0457} &[0.0205, 0.0708] &  \textcolor{blue}{0.1148} &[0.1023, 0.1272]&  \textcolor{blue}{0.1039}&[0.0947, 0.1132]& -&- \\
				& \textbf{Ours} &  \textcolor{red}{0.0753} &[0.0702, 0.0805]& \textcolor{red}{0.0269} &[0.0181, 0.0356] &  \textcolor{red}{0.0845} &[0.0768, 0.0922]&  \textcolor{red}{0.1015}&[0.0942, 0.1087]& -&- \\
				\hline
				\multirow{4}{*}{Stomach-III} 
				& Endo-SfM & 0.1993 &[0.1786, 0.2200]&  \textcolor{blue}{0.1622}&[0.1480, 0.1764] & 0.1978&[0.1798, 0.2159]& 0.5093 &[0.4645, 0.5541]& -&- \\
				& ManyDepth & 0.1930 &[0.1737, 0.2124]& 0.1851 &[0.1668, 0.2033] & 0.1813 &[0.1647, 0.1979]& 0.4765&[0.4347, 0.5182]& -&- \\
				& AF-SfMLearner &  \textcolor{blue}{0.1558} &[0.1398, 0.1718]& 0.1750&[0.1581, 0.1918] &  \textcolor{blue}{0.1497}&[0.1386, 0.1607]&  \textcolor{blue}{0.4358}&[0.3940, 0.4777]& -&- \\
				& \textbf{Ours} &  \textcolor{red}{0.1085} &[0.0978, 0.1192]& \textcolor{red}{0.1486}&[0.1341, 0.1631] &  \textcolor{red}{0.1084}&[0.1006, 0.1163]&  \textcolor{red}{0.2960}&[0.2634, 0.3286]& -&- \\
				\hline	
		\end{tabular}}
	\end{center}
	\label{table10}
\end{table*}

\begin{table}[htb!]
	\caption{Quantitative pose comparison averaged over each organ on the EndoSLAM dataset. }
	\begin{center}
		\renewcommand{\arraystretch}{1.3}
		\resizebox{0.88\columnwidth}{!}{\begin{tabular}{c c c c }
				\hline
				&\multirow{2}{*}{Method} & \multicolumn{2}{c}{Trajectories} \\ 
				\cline{3-4} 
				&& ATE ($\times$ 1e-2)& 95\% CIs    \\ 
				\hline
				\hline
				\multirow{3}{*}{Colon-IV} 
				& Endo-SfM~\citep{bengisu2020quantitative} & 0.1096 &[0.0884, 0.1307]\\
				& ManyDepth~\cite{watson2021temporal} & 0.1127 &[0.0985, 0.1269]\\
				& AF-SfMLearner~\citep{shao2021selfsupervised} & \textcolor{blue}{0.0917} &[0.0811, 0.1023]\\
				& \textbf{Ours} & \textcolor{red}{0.0844} &[0.0739, 0.0949]\\
				\hline
				\multirow{3}{*}{Intestine} 
				& Endo-SfM~\citep{bengisu2020quantitative} & 0.1690 &[0.1597, 0.1783] \\
				& ManyDepth~\cite{watson2021temporal} & 0.1835 &[0.1738, 0.1933]\\
				& AF-SfMLearner~\citep{shao2021selfsupervised} & \textcolor{blue}{0.1133} &[0.1082, 0.1184]\\
				& \textbf{Ours} & \textcolor{red}{0.1058} &[0.1013, 0.1102]\\
				\hline
				\multirow{3}{*}{Stomach-I}
				& Endo-SfM~\citep{bengisu2020quantitative}  & 0.1790 &[0.1586, 0.1994] \\
				& ManyDepth~\cite{watson2021temporal} & 0.2012 &[0.1788, 0.2236]\\
				& AF-SfMLearner~\citep{shao2021selfsupervised}  & \textcolor{blue}{0.1662} &[0.1449, 0.1874]\\	
				& \textbf{Ours} & \textcolor{red}{0.1481} &[0.1299, 0.1662]\\
				\hline
				\multirow{3}{*}{Stomach-II} 
				& Endo-SfM~\citep{bengisu2020quantitative} & 0.1158 &[0.1091, 0.1225] \\
				& ManyDepth~\cite{watson2021temporal} & 0.1158 &[0.1093, 0.1224]\\
				& AF-SfMLearner~\citep{shao2021selfsupervised}  & \textcolor{blue}{0.1035} &[0.0976, 0.1094]\\
				& \textbf{Ours} & \textcolor{red}{0.0869} &[0.0829, 0.0909]\\
				\hline
				\multirow{3}{*}{Stomach-III} 
				& Endo-SfM~\citep{bengisu2020quantitative} & 0.2690 &[0.2545, 0.2835]\\
				& ManyDepth~\cite{watson2021temporal} & 0.2602 &[0.2465, 0.2740]\\
				& AF-SfMLearner~\citep{shao2021selfsupervised}  & \textcolor{blue}{0.2301} &[0.2172, 0.2431]\\
				& \textbf{Ours} & \textcolor{red}{0.1659} &[0.1560, 0.1758]\\
				\hline	
		\end{tabular}}
	\end{center}
	\label{table11}
\end{table}

To validate the efficacy of the proposed model, we perform extensive experiments on three tasks: depth estimation, pose estimation and point cloud estimation. For depth estimation, we provide results on the SCARED dataset, Hamlyn dataset and SERV-CT dataset, as well as qualitative results on the challenging real part of EndoSLAM dataset. We use the split of~\citep{shao2021selfsupervised} on the SCARED dataset, which is composed of 15351, 1705 and 551 frames for training set, validation set and test set, respectively. In addition, we randomly choose 250 frames as the test set on the Hamlyn dataset from~\citep{recasens2021endo}.  We conduct pose comparison on the SCARED and EndoSLAM datasets. We choose the videos from dataset3/keyframe4 and dataset5/keyframe4 following~\citep{shao2021selfsupervised} on the SCARED dataset. Besides, we perform 5-fold cross-validation on the LowCam of the real part on the EndoSLAM dataset, where each of the 5 models is evaluated on videos from one organ and trained on the rest. For point cloud estimation, we compare the methods on the SCARED dataset.


\subsection{Experimental settings}

The proposed framework is implemented using the PyTorch library~\citep{paszke2017automatic}. We optimize it with the Adam optimizer~\citep{kingma2014adam}, where ${\beta _1} = 0.9,{\beta _2} = 0.99$ and a batch size of 12. The total number of epochs is set to 20 and the initial learning rate is ${{10}^{-4}}$, dropping by a factor of 10 after 10 epochs. The input resolution is $320 \times 256$ pixels. As with existing works~\citep{bengisu2020quantitative,shao2021selfsupervised,watson2021temporal}, we use weights pretrained on ImageNet~\citep{deng2009imagenet}. Besides, we perform static cameras and start of sequences augmentations the same as~\citep{watson2021temporal}, and use the standard evaluation metrics based on~\citep{shao2021selfsupervised}.



During the evaluation phase, we apply the median scaling technique~\citep{zhou2017unsupervised} to the predicted depth maps, written as
\begin{equation}
	\resizebox{0.85\hsize}{!}{$
		{\bm{{\rm{D}}}_{scaled}}\! =\! \left( {{\bm{{\rm{D}}}_{pred}}\! *\! \left( {{{median\left( {{\bm{{\rm{D}}}_{gt}}} \right)} \mathord{\left/
						{\vphantom {{median\left( {{\bm{{\rm{D}}}_{gt}}} \right)} {median\left( {{\bm{{\rm{D}}}_{pred}}} \right)}}} \right.
						\kern-\nulldelimiterspace} {median\left( {{\bm{{\rm{D}}}_{pred}}} \right)}}} \right)} \right)$}.
\end{equation}
Following~\citep{shao2021selfsupervised}, we clip the scaled depth maps at 150 mm on the SCARED and Hamlyn datasets, and 180 mm on the SERV-CT dataset.

\begin{figure*}[!htb]
	\centering
	\includegraphics[width=0.9\linewidth]{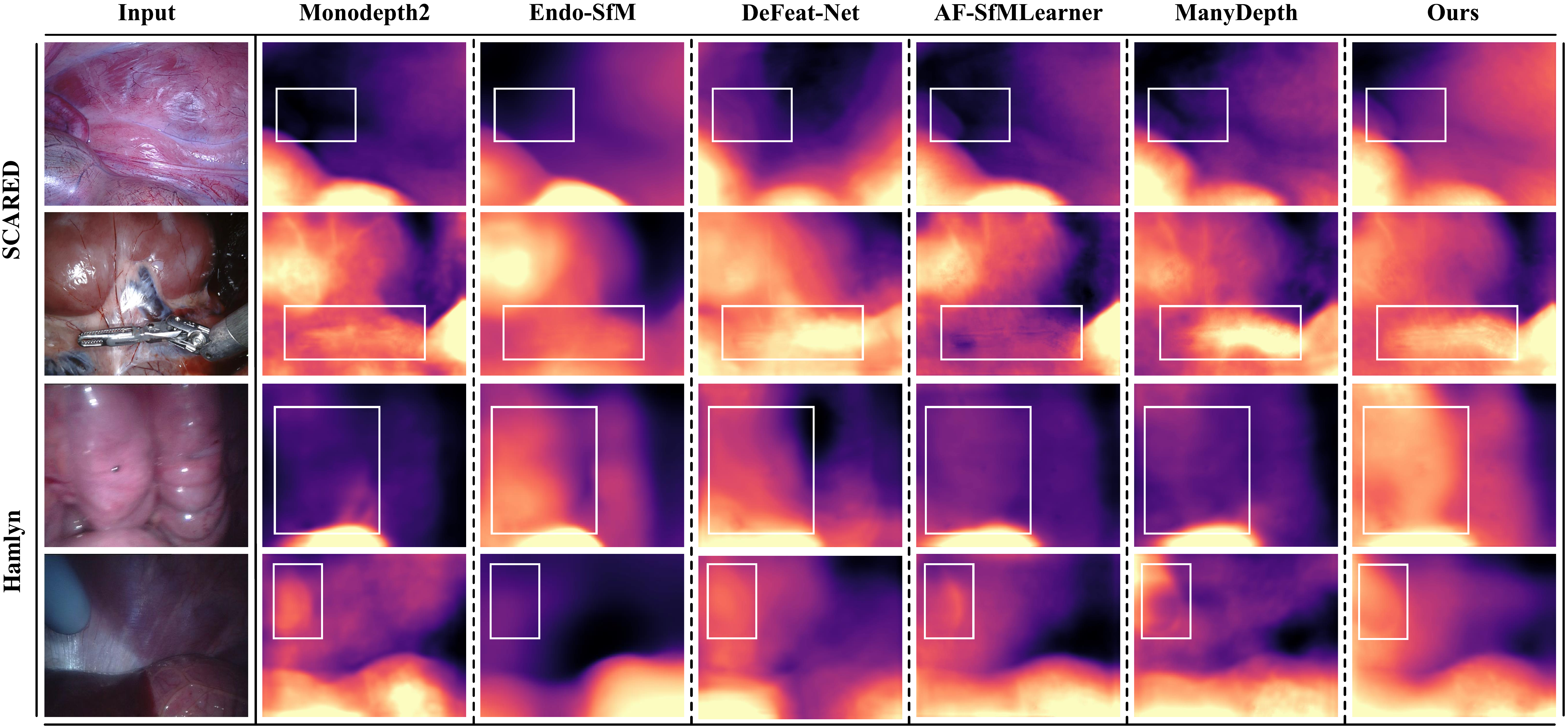}
	\caption{Qualitative depth comparison of our method against the compared methods on the SCARED and Hamlyn datasets. The white boxes highlight the regions to focus on.} 
\label{scared}
\end{figure*}

\begin{figure*}[!htb]
\centering
\includegraphics[width=0.9\linewidth]{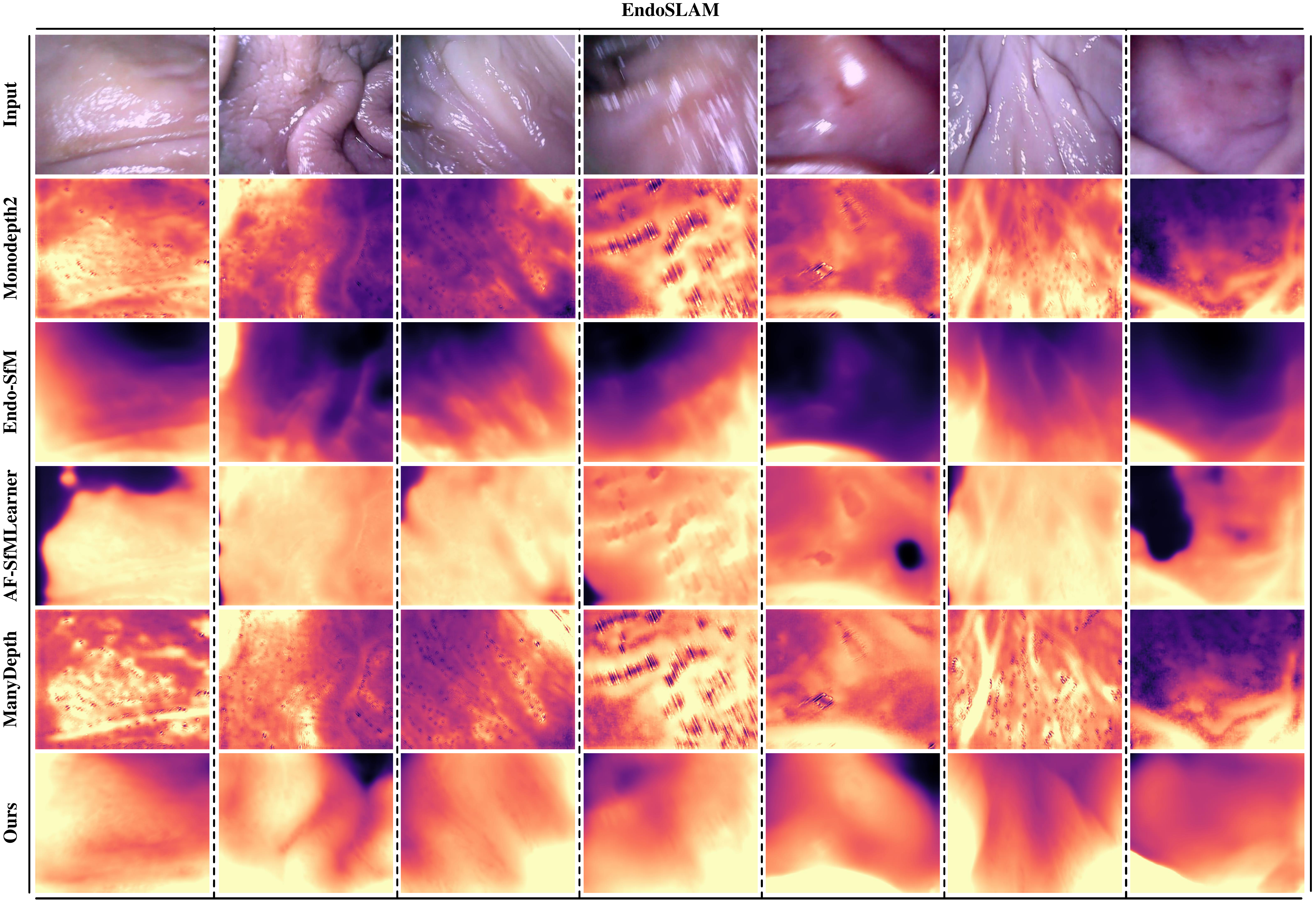}
\caption{Qualitative depth comparison on the EndoSLAM dataset.}	
\label{endoslam}
\end{figure*}
\subsection{Comparison to state-of-the-art competitors}
The results of compared single-frame monocular methods are  from~\citep{shao2021selfsupervised} and~\cite{yang2024self1}. For ManyDepth \citep{watson2021temporal}, we reproduce the algorithm with the official implementations and evaluate it on the corresponding datasets. 
\subsubsection{Depth estimation} Table~\ref{table1} demonstrates the depth comparison results on the SCARED dataset. As can be seen, our model outperforms the compared methods by a large margin. In particular, it improves the AF-SfMLearner and ManyDepth by more than 10$\%$ and 15$\%$ on most metrics, respectively. Besides, the results of multi-frame monocular methods are generally better than those of single-frame monocular methods, which reveals that the additional input frames are valuable and contribute to the performance gain. It is worth noting that while our model and ManyDepth both adopt only two frames to construct the cost volume during the training phase, when more frames are put into the cost volume building at evaluation time, our model can improve the accuracy, whereas ManyDepth fails. 

Fig.~\ref{scared} shows the qualitative comparison on the SCARED dataset and Hamlyn dataset, which indicates that our method is better at textureless regions, such as finger. Besides, we notice that most methods are vulnerable to the texture details, which lead to slight depth variations in these regions, while the Endo-SfM performs more robustly. It may be induced by the inability to fully exploit the contextual information owing to the limited receptive field of the CNN. Although the Endo-SfM is also built upon CNN, a large weight is imposed on smoothness loss to enforce depth consistency in Endo-SfM, but also comes with over-smooth artifacts.
\begin{figure*}[!htb]
	\centering
	\includegraphics[width=0.88\linewidth]{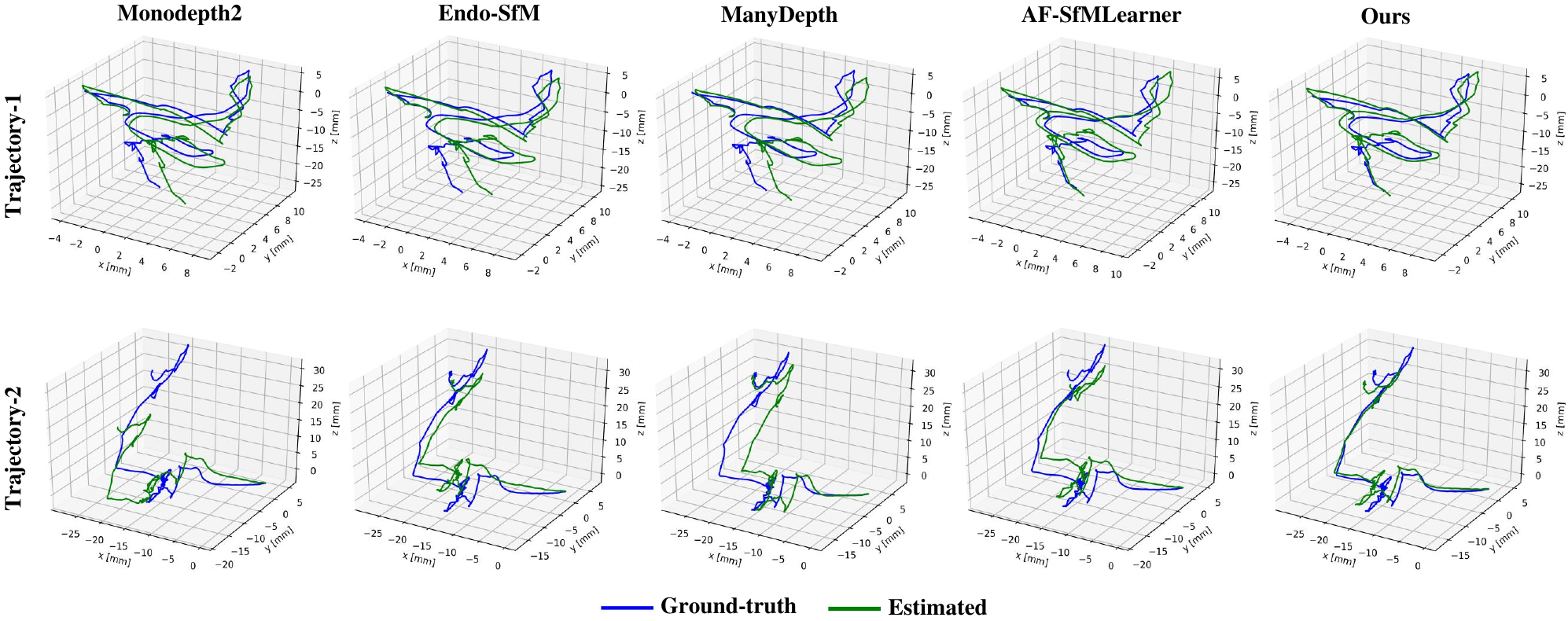}
	\caption{Qualitative pose comparison on the trajectory-1 (410 frames) and trajectory-2 (833 frames), which are from the SCARED dataset. Following~\citep{shao2021selfsupervised}, we calculate a global scale over the whole sequence to recover the absolute scale. }
	\label{Fig17}
\end{figure*}

\begin{table*}[htb!]
	\caption{Hamlyn depth results (in vivo videos). All models are unsupervised trained on the SCARED dataset.}
	\begin{center}
		\renewcommand{\arraystretch}{1.3}
		\resizebox{1.89\columnwidth}{!}{\begin{tabular}{c c c c c c c c c c c c}	
				\Xhline{1.2pt}
				Method & V& \cellcolor {orange!20} Abs Rel $\downarrow$ & 95\% CIs & \cellcolor {orange!20} Sq Rel $\downarrow$ & 95\% CIs & \cellcolor {orange!20} RMSE $\downarrow$& 95\% CIs& \cellcolor {orange!20} RMSE log $\downarrow$ &95\% CIs & \cellcolor {blue!20} $\delta$ $\uparrow$ & 95\% CIs\\ 
				\hline						
				\hline
				SfMLearner~\citep{zhou2017unsupervised} &1& 0.152 & [0.142, 0.162] & 2.582 & [2.151, 3.012] & 10.481 & [9.418, 11.544] & 0.205 & [0.194, 0.217]& 0.761 & [0.737, 0.785]\\
				DeFeat-Net~\citep{spencer2020defeat} &1& \textcolor{blue}{0.143} & [0.132, 0.154] & 2.663 & [2.136, 3.191] & 10.323 & [9.155, 11.491]& 0.191 & [0.179, 0.204] & \textcolor{blue}{0.797} & [0.772, 0.822] \\
				Monodepth2~\citep{godard2019digging} &1& 0.167 & [0.155, 0.179] & 3.383 & [2.728, 4.037]& 11.239& [10.000, 12.477] & 0.205 & [0.193, 0.218] & 0.758 & [0.732, 0.783]\\
				Endo-SfM~\citep{bengisu2020quantitative}&1&  0.147 &[0.135, 0.158] & 2.800 & [2.234, 3.366] & 10.207 & [9.025, 11.390]& \textcolor{blue}{0.188} & [0.175, 0.200]& 0.795 & [0.768, 0.821]\\
				AF-SfMLearner~\citep{shao2021selfsupervised} &1& \textcolor{blue}{0.143}& [0.132, 0.154]& \textcolor{blue}{2.561} & [2.066, 3.055]& \textcolor{blue}{10.065}& [8.971, 11.160]& \textcolor{blue}{0.188} & [0.177, 0.200]& 0.796& [0.772, 0.821]\\
				\hline
				\hline
				ManyDepth~\cite{watson2021temporal} & 2 & 0.155 &[0.145, 0.165] &2.743 &[2.228, 3.257] & 10.367 &[9.209, 11.524]& 0.194 &[0.182, 0.206]&0.781&[0.756, 0.806]
				\\
				\textbf{Ours} &2 & \textcolor{red}{0.138}& [0.128, 0.148]&\textcolor{red}{2.316} &[1.840, 2.792] & \textcolor{red}{9.351} &[8.286, 10.416]&\textcolor{red}{0.175} &[0.164, 0.186]&\textcolor{red}{0.810}&[0.785, 0.836]
				\\
				\Xhline{1.2pt}
		\end{tabular}}
	\end{center}
	\label{table2}
\end{table*}

\begin{table*}[htb!]
	\caption{SERV-CT depth results. Because the SERV-CT dataset does not contain video but stereo pairs, we adopt the stereo pairs as the input for multi-frame monocular methods.}
	\begin{center}
		\renewcommand{\arraystretch}{1.3}
		\resizebox{1.9\columnwidth}{!}{\begin{tabular}{c c c c c c c c c c c c}	
				\Xhline{1.2pt}
				Method & V& \cellcolor {orange!20} Abs Rel $\downarrow$ & 95\% CIs & \cellcolor {orange!20} Sq Rel $\downarrow$ & 95\% CIs & \cellcolor {orange!20} RMSE $\downarrow$& 95\% CIs& \cellcolor {orange!20} RMSE log $\downarrow$ &95\% CIs & \cellcolor {blue!20} $\delta$ $\uparrow$ & 95\% CIs\\ 
				\hline						
				\hline
				SfMLearner~\citep{zhou2017unsupervised} &1& 0.151 & [0.137, 0.165] & 3.917  & [3.155, 4.680] & 17.451 & [15.298, 19.604] & 0.191 & [0.176, 0.207]& 0.779 & [0.741, 0.817]\\
				DeFeat-Net~\citep{spencer2020defeat} &1& 0.114 & [0.105, 0.124] &  1.946 & [1.534, 2.358] & 12.588 & [10.888, 14.287]& 0.153 & [0.139, 0.166] &  0.873 & [0.843, 0.902] \\
				Monodepth2~\citep{godard2019digging} &1& 0.123 & [0.112, 0.134] & 2.205 & [1.727, 2.683]& 12.927& [11.296, 14.557] & 0.152 & [0.138, 0.165] & 0.856 & [0.824, 0.888]\\
				Endo-SfM~\citep{bengisu2020quantitative}&1 & 0.116 &[0.105, 0.127] & 2.014 & [1.610, 2.419] & 12.493 & [10.921, 14.066]&  0.143 & [0.130, 0.157]& 0.864 & [0.829, 0.899]\\
				AF-SfMLearner~\citep{shao2021selfsupervised} &1& \textcolor{blue}{0.102}& [0.091, 0.113]& \textcolor{blue}{1.632} & [1.235, 2.029]& \textcolor{blue}{11.092}& [9.432, 12.751]& \textcolor{blue}{0.131} & [0.116, 0.145]& \textcolor{blue}{0.898}& [0.868, 0.929]\\
				\hline
				\hline
				ManyDepth~\cite{watson2021temporal} & 2 & 0.117 &[0.106, 0.128] &1.922 &[1.509, 2.335] & 12.128 &[10.518, 13.737]& 0.146 &[0.134, 0.159]&0.868&[0.837, 0.898]
				\\
				\textbf{Ours} &2 & \textcolor{red}{0.098}& [0.092, 0.104]&\textcolor{red}{1.402} &[1.178, 1.626] & \textcolor{red}{10.656} &[9.417, 11.896]&\textcolor{red}{0.126} &[0.118, 0.134]&\textcolor{red}{0.916}&[0.898, 0.934]
				\\
				\Xhline{1.2pt}
		\end{tabular}}
	\end{center}
	\label{table8}
\end{table*}
\begin{table}[htb!]
	\caption{Ablation study on the entire framework. CV: cost volume; LPM: learnable patchmatch module; CT: cross-teaching; ST: self-teaching. }
	\begin{center}
		\renewcommand{\arraystretch}{1.3}
		\resizebox{0.95\columnwidth}{!}{\begin{tabular}{c c c c c c c c c}
				\Xhline{1.2pt}
				ID & CV & LPM & CT & ST& \cellcolor {orange!20} Abs Rel $\downarrow$ & \cellcolor {orange!20} Sq Rel $\downarrow$ & \cellcolor {orange!20} RMSE $\downarrow$ & \cellcolor {orange!20} RMSE log$\downarrow$\\ 
				\hline
				\hline				
				1&&& && 0.072 & 0.635 & 6.047 & 0.099\\
				2 &\cmark&&&& 0.068 & 0.560 & 5.494 & 0.092\\	
				3 &\cmark&\cmark&&& 0.059 & 0.443 & 4.996 & 0.082\\	
				4 &\cmark&\cmark&\cmark&& 0.055 & 0.410 & 4.825 & 0.078\\
				5 &\cmark&\cmark&\cmark&\cmark& \textcolor{red}{0.051}& \textcolor{red}{0.386} & \textcolor{red}{4.675} & \textcolor{red}{0.074} \\
				\Xhline{1.2pt}			
		\end{tabular}}
	\end{center}
	\label{table3}
\end{table}

Fig.~\ref{endoslam} demonstrates the qualitative comparison on the EndoSLAM dataset. The EndoSLAM is full of severe inter-frame brightness changes and textureless regions. It can be seen that our model generates more continuous depth maps and better delineates anatomical structures. Besides, the Monodepth2 and ManyDepth are susceptible to the over-saturated regions. This is due to the fact that the boundary parts of these over-saturated regions are more prone to brightness fluctuations, resulting in ambiguous supervision of photometric loss.


\subsubsection{Pose estimation}
Table~\ref{table7} presents the pose comparison results on the SCARED dataset, and our model achievers lower ATE on both trajectories. Since the depth estimation and the pose estimation are tightly coupled during the training phase, improved depth leads to improvements in the pose as well. Table~\ref{table10} and Table~\ref{table11} present the quantitative results on the challenging EndoSLAM dataset. As can be seen from Table~\ref{table10}, our model is able to exceed the compared methods for most of the cases. In Table~\ref{table11}, we further list the performance averaged on each organ. The superior results in both tables validate the efficacy of the proposed components. Additionally, we notice that even for folds that do not contain colon or small intestine during training, our model achieves low ATE in these two organs, which is an indicator of its excellent organ adaptability.

Fig.~\ref{Fig17} displays the predicted trajectory-1 and trajectory-2 on the SCARED dataset. It can be seen that the trajectories of ours are generally better than those of the compared methods, especially for the longer trajectory-2.

\subsubsection{Point cloud estimation} We generate point clouds using the depth predictions and camera intrinsic. To overcome the inherent scale ambiguity, we adopt the same median scaling strategy as in depth evaluation. Table~\ref{table6} summarizes the point cloud comparison results on the SCARED dataset, which indicates that our model surpasses the compared methods, including both single-frame monocular and multi-frame monocular. The SfMLearner, DeFeat-Net, Monodepth2 and ManyDepth are heavily affected by the brightness fluctuations and low and homogeneous textures. While the AF-SfMLearner and Endo-SfM are specifically designed for endoscopic scenarios, they cannot enhance their depth predictions by leveraging valuable additional input frames. In contrast, our model could benefit from both.

	\subsection{Generalization to the Hamlyn and SERV-CT datasets}
	We investigate the generalization ability to verify that our model indeed learns transferable features, rather than simply memorizing training samples. To this end, we apply the models trained on SCARED to the Hamlyn and SERV-CT datasets for evaluation without any fine-tuning. The frames are resized to $320 \times 256$ pixels. As presented in Table~\ref{table2} and Table~\ref{table8}, the superior results indicate that our model generalizes well across different patients and cameras. It is worth mentioning that although the stereo pairs used on the SERV-CT dataset have significantly different relative poses from the video frames for training, our model achieves good results (Table~\ref{table8}). 
	\begin{figure*}[!htb]
		\centering
		\includegraphics[width=0.9\linewidth]{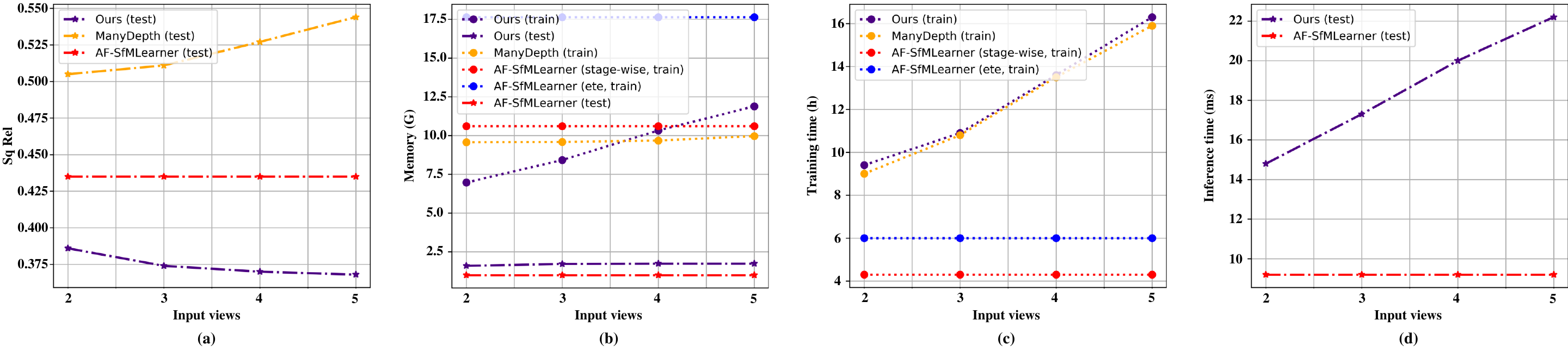}
		\caption{Comparison of multi-frame monocular methods with the single-frame monocular method AF-SfMLearner on Sq Rel (a), memory (b), training time (c) and inference time (d). In (a), ManyDepth and ours are trained with an input view of 2 during training. Since ManyDepth has almost the identical inference memory and inference time as ours, we do not show the relevant results of ManyDepth in (b) and (d).  AF-SfMLearner (stage-wise, train): the model is trained in a stage-wise manner. AF-SfMLearner (ete, train): the model is trained in an end-to-end manner. The AF-SfMLearner (stage-wise, train) and ours (train) only include the time of training the DepthNet stage on training time.}
		\label{smti}
	\end{figure*}
		\begin{figure}[!htb]
		\centering
		\includegraphics[width=0.9\linewidth]{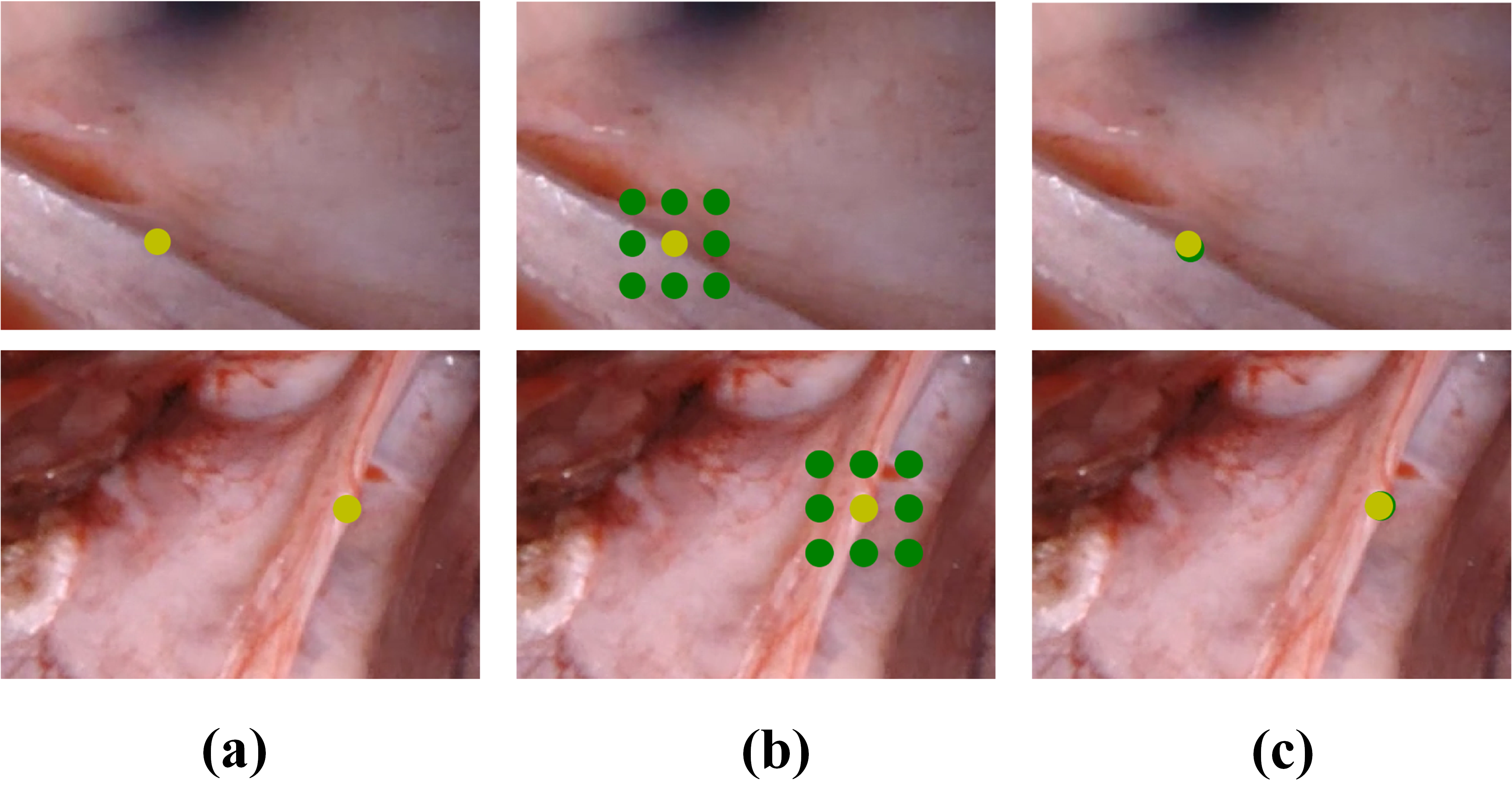}
		\caption{Visualization of the sampled locations in sharp edge regions. (a) Target image. (b) Sampled locations of patchmatch module~\citep{yu2020p}. (c) Sampled locations of our learnable patchmatch module. The yellow marker and green marker denote the center pixel and sampled pixel, respectively. }	
		\label{Fig4}
	\end{figure}
	
	\begin{figure}[!htb]
		\centering
		\includegraphics[width=0.91\linewidth]{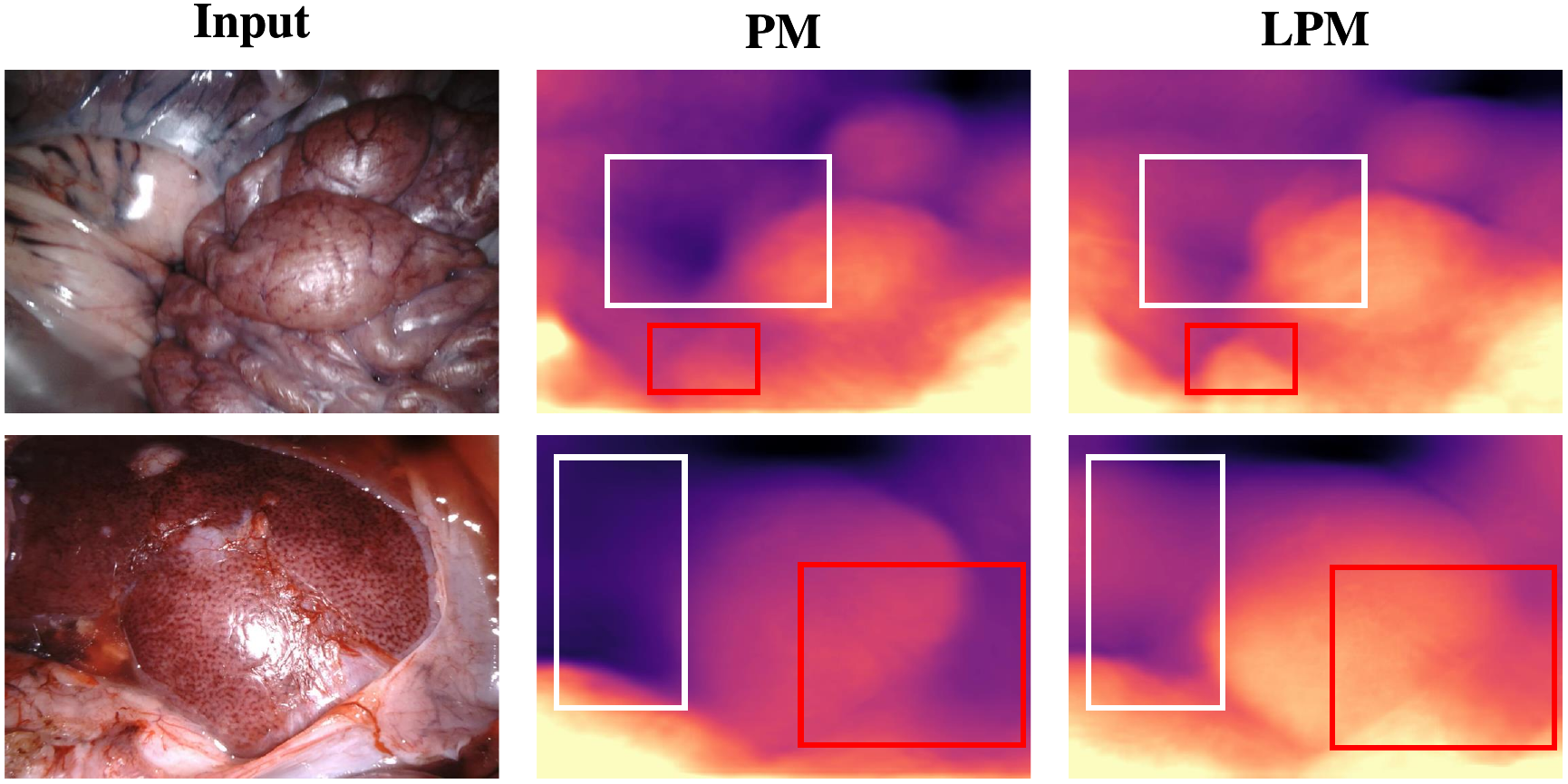}
		\caption{Qualitative depth results for ID 4 (patchmatch module~\citep{yu2020p}) and ID 5 (learnable patchmatch module) in Table~\ref{table5}. The white and red boxes indicate the regions to focus on.}	
		\label{Fig16}
	\end{figure}
	\begin{figure}[!htb]
		\centering
		\includegraphics[width=1.0\linewidth]{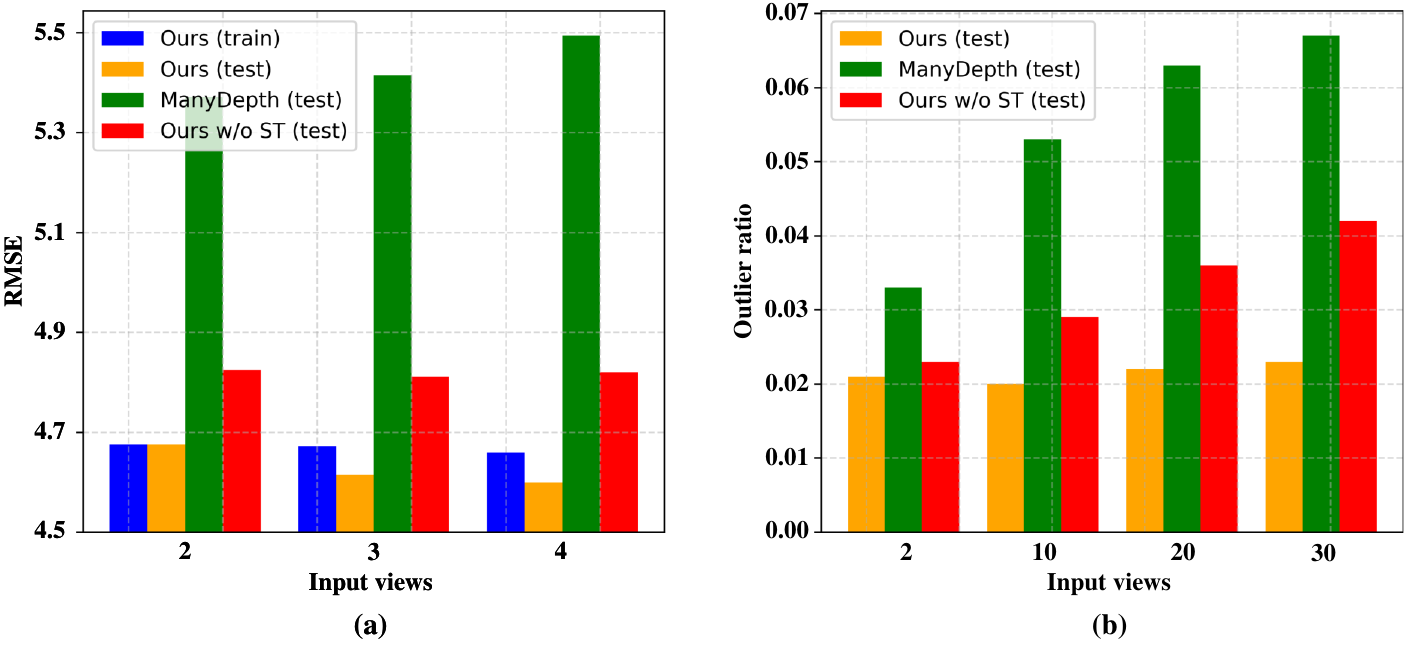}
		\caption{Ablation study on the self-teaching paradigm. (a) RMSE. (b) Outlier ratio, which is computed by $1-\delta$.}	
		\label{Fig15}
	\end{figure}

	\subsection{Ablation studies}
	We conduct several ablation studies, including entire framework, break-down comparison of learnable patchmatch module, self-teaching paradigm and multi-frame vs single-frame methods. 
	
	\begin{table}[htb!]
		\caption{Break-down comparison of learnable patchmatch module. SAM: spatial attention module~\citep{bengisu2020quantitative}; FSM: feature scaling module~\citep{shao2021self}; PM~\citep{yu2020p}: patchmatch module; LPM: learnable patchmatch module.}
		\begin{center}
			\renewcommand{\arraystretch}{1.3}
			\resizebox{1.0\columnwidth}{!}{\begin{tabular}{c c c c c c c c c}
					\Xhline{1.2pt}
					ID & SAM& FSM& PM& LPM & \cellcolor {orange!20} Abs Rel $\downarrow$ &  \cellcolor {orange!20} Sq Rel $\downarrow$ & \cellcolor {orange!20} RMSE $\downarrow$ & \cellcolor {orange!20} RMSE log $\downarrow$ \\ 
					\hline
					\hline				
					1&&& && 0.072 & 0.635 & 6.047 & 0.099 \\
					2 &\cmark&&&& 0.070 & 0.582 & 5.818 & 0.095 \\	
					3 &&\cmark&&& 0.069 & 0.562 & 5.620 & 0.093 \\
					4 &&&\cmark&& 0.066 & 0.532 & 5.414 & 0.092 \\
					5 &&&&\cmark& \textcolor{red}{0.063} & \textcolor{red}{0.487} & \textcolor{red}{5.183} & \textcolor{red}{0.086} \\
					\Xhline{1.2pt}				
			\end{tabular}}
		\end{center}
		\label{table5}
	\end{table}

	\subsubsection{Entire framework (Table~\ref{table3})}  We start with the underperforming baseline (ID 1). By first appending the cost volume, we observe a slight improvement in performance (ID 2). ID 3 presents an addition of the learnable patch module and shows considerable efficacy in boosting performance. We then add the cross-teaching paradigm, again with a consistent improvement on all evaluation metrics (ID 4). We finally integrate the self-teaching paradigm and form the entire framework, which achieves the best results (ID 5).


	\subsubsection{Break-down comparison of learnable patchmatch module (Table~\ref{table5})}
	To further show the superiority of learnable patchmatch module, we perform comparison with other alternative methods, including spatial attention module~\citep{bengisu2020quantitative}, patchmatch module~\citep{yu2020p}, and feature scaling module~\citep{shao2021self}. It can be seen that our learnable patchmatch module outperforms other alternative methods. Compared with the patchmatch module, our module can avoid severe errors by gathering pixels of the same depth and improves it by approximately 8.5$\%$ on the Sq Rel that is sensitive to the large depth errors (IDs 4 and 5). 
	Fig.~\ref{Fig4} displays the sampled locations of patchmatch module and our learnable patchmatch module in the sharp edge regions. Ours is able to adaptively aggregate pixels near the center pixel in this case. Fig.~\ref{Fig16} presents the qualitative depth comparison of patchmatch module and learnable patchmatch module. The model with the patchmatch module is prone to errors due to depth changes while ours preserves details such as boundaries.
	
	\subsubsection{Self-teaching paradigm (Fig.~\ref{Fig15})} 
	Ours (test), Ours w/o ST (test) and ManyDepth (test) have two input frames during the training phase. It can be seen that when increasing the input frames at training time, the performance gain is small, which is consistent with the finding in ManyDepth. However, unlike ManyDepth, our model can improve the depth accuracy when increasing input frames at test time. To figure out why, we further present the depth accuracy after ablating the self-teaching paradigm and find that it roughly matches the case of ManyDepth, which reveals that the self-teaching paradigm enables model to leverage more frames to improve the depth predictions at test time, although only two input frames are utilized during training. This is not surprising given that the self-teaching paradigm helps teach model to ignore the detrimental information and focus on the meaningful elements in the cost volume. As the number of input frames increases, so do the occlusions and brightness fluctuations between source and target frames, which are disasters for other methods but not for our model equipped with the self-teaching. Furthermore, we increase the number of input frames to 10, 20 or even 30, at which time the cost volume is almost overwhelmed by noise, and our model only generates few outliers, indicating that it has high immunity against input noise in the cost volume. This is a desired property for model in practical applications because the data may contain heavy noise.
	
	\subsubsection{Multi-frame vs single-frame methods (Fig.~\ref{smti})} To know the relationship between cost and performance gain brought by multiple frames, we compare the multi-frame monocular methods with the single-frame monocular method AF-SfMLearner on Sq Rel, training/inference memory and training/inference time. It can be seen that the training time and training memory of our model increase rapidly with the number of input frames, while the performance gain is limited as shown in Fig.~\ref{Fig15}(a). Therefore, we take two input frames during the training phase, in which case our model achieves less training memory than ManyDepth and AF-SfMLearner. During the evaluation phase, the performance of our model improves more than the training phase as the number of input frames increases, while the cost is relatively small, for instance, inference memory. Although the inference time increases with input frames, our model can still achieve about 45 frames per second at a resolution of $320 \times 256$ pixels when the number of input frames is 5, which meets the real-time requirements in most applications.

	\section{Conclusion}
	In this work, we present a novel unsupervised model, which predicts superior depths from multiple endoscopic frames. On the one hand, the model incorporates a learnable patchmatch module to adaptively increase the discriminative capability in low- and homogeneous-texture regions. On the other hand, the model enforces cross-teaching and self-teaching consistencies to provide efficacious regularizations towards brightness variations. When more frames are put into the cost volume building at test time, the model can improve the depth predictions. This enables us to flexibly adjust the number of input frames in light of the accuracy requirement. 
	Extensive experiments on four datasets including SCARED, EndoSLAM, Hamlyn and SERV-CT validate the efficacy of the model. 
	
	
	\normalem
	\section{Acknowledgement}
	This work was supported by the Key Research and Development Program of Shandong Province under Grant No. 2019JZZY011101.
	

\bibliographystyle{cas-model2-names}

\bibliography{refers}



\end{document}